\documentclass[lettersize,journal]{IEEEtran}
\usepackage{amsmath,amsfonts}
\usepackage[table,xcdraw]{xcolor}
\usepackage{algorithm}
\usepackage{algorithmicx}
\usepackage{algpseudocode}
\algloopdefx{NoEndIf}[1]{\textbf{If} #1 \textbf{then}}
\algloopdefx{NoEndElif}[1]{\textbf{Elif} #1 \textbf{then}}
\algloopdefx{NoEndFor}[1]{\textbf{For} #1 \textbf{do}}
\algloopdefx{NoEndWhile}[1]{\textbf{While} #1 \textbf{do}}
\usepackage{amsmath}
\usepackage{array}
\usepackage[caption=false,font=normalsize,labelfont=sf,textfont=sf]{subfig}
\usepackage{textcomp}
\usepackage{stfloats}
\usepackage{url}
\usepackage{verbatim}
\usepackage{graphicx}
\usepackage{cite}
\usepackage{scalerel}
\usepackage{tikz}

\usetikzlibrary{calc}
\makeatletter
\def\mfontsize{\f@size}

\makeatother
\definecolor{darkpurple}{rgb}{0.6, 0.3, 0.8}  % Adjust the RGB values to get the desired shade

\usepackage{comment}
\usetikzlibrary{svg.path}
\definecolor{orcidlogocol}{HTML}{A6CE39}
\tikzset{
  orcidlogo/.pic={
    \fill[orcidlogocol] svg{M256,128c0,70.7-57.3,128-128,128C57.3,256,0,198.7,0,128C0,57.3,57.3,0,128,0C198.7,0,256,57.3,256,128z};
    \fill[white] svg{M86.3,186.2H70.9V79.1h15.4v48.4V186.2z}
                 svg{M108.9,79.1h41.6c39.6,0,57,28.3,57,53.6c0,27.5-21.5,53.6-56.8,53.6h-41.8V79.1z M124.3,172.4h24.5c34.9,0,42.9-26.5,42.9-39.7c0-21.5-13.7-39.7-43.7-39.7h-23.7V172.4z}
                 svg{M88.7,56.8c0,5.5-4.5,10.1-10.1,10.1c-5.6,0-10.1-4.6-10.1-10.1c0-5.6,4.5-10.1,10.1-10.1C84.2,46.7,88.7,51.3,88.7,56.8z};
  }
}

\newcommand\orcidicon[1]{\href{https://orcid.org/#1}{\mbox{\scalerel*{
\begin{tikzpicture}[yscale=-1,transform shape]
\pic{orcidlogo};
\end{tikzpicture}
}{|}}}}
\usepackage{hyperref} 

\usepackage{multirow}

\usepackage{graphicx}
\usepackage{algpseudocode}

\usepackage{enumitem}
\newlist{questions}{enumerate}{2}
\setlist[questions,1]{label=RQ\arabic*.,ref=RQ\arabic*}
\setlist[questions,2]{label=(\alph*),ref=\thequestionsi(\alph*)}

\begin{document}

\title{Generating Realistic Adversarial Examples for Business Processes using Variational Autoencoders}
\author{\IEEEauthorblockN{Alexander Stevens\orcidicon{0000-0001-6140-8788}, Jari Peeperkorn\orcidicon{0000-0003-4644-4881}, Johannes De Smedt\orcidicon{0000-0003-0389-0275},
Jochen De Weerdt\orcidicon{0000-0001-6151-0504}}\\
\IEEEauthorblockA{\textit{Research Centre for
Information Systems
Engineering (LIRIS)} \\
\textit{KU Leuven},\\ Leuven, Belgium}}

% The paper headers
\markboth{IEEE Transactions on Knowledge and Data Engineering}%
{Shell \MakeLowercase{\textit{et al.}}: A Sample Article Using IEEEtran.cls for IEEE Journals}

\IEEEpubid{0000--0000/00\$00.00~\copyright~2021 IEEE}

% Remember, if you use this you must call \IEEEpubidadjcol in the second
% column for its text to clear the IEEEpubid mark.

\maketitle

\begin{abstract}
In predictive process monitoring, predictive models are vulnerable to adversarial attacks, where input perturbations can lead to incorrect predictions. Unlike in computer vision, where these perturbations are designed to be imperceptible to the human eye, the generation of adversarial examples in predictive process monitoring poses unique challenges. Minor changes to the activity sequences can create improbable or even impossible scenarios to occur due to underlying constraints such as regulatory rules or process constraints. To address this, we focus on generating \emph{realistic} adversarial examples tailored to the business process context, in contrast to the imperceptible, pixel-level changes commonly seen in computer vision adversarial attacks. This paper introduces two novel \emph{latent space attacks}, which generate adversaries by adding noise to the latent space representation of the input data, rather than directly modifying the input attributes. These latent space methods are domain-agnostic and do not rely on process-specific knowledge, as we restrict the generation of adversarial examples to the learned class-specific data distributions by directly perturbing the latent space representation of the business process executions.

We evaluate these two latent space methods with six other adversarial attacking methods on eleven real-life event logs and four predictive models. The first three attacking methods directly permute the activities of the historically observed business process executions. The fourth method constrains the adversarial examples to lie within the same data distribution as the original instances, by \emph{projecting} the adversarial examples to the original data distribution. 
\end{abstract}


\section{Introduction}\label{sec:introduction}
\IEEEPARstart{M}{achine} Learning (ML) and Deep Learning (DL) methods have become central to a wide range of applications such as loan application processes \cite{babaev2019rnn}, criminal justice \cite{rudin2020age} and, disease diagnostics \cite{DBLP:journals/natmi/Rudin19}. Similarly, there has been a growing interest in using ML and DL techniques within Process Mining (PM), a field focused on analyzing event data generated during the execution of business processes. PM enables organizations to uncover, monitor, and improve real-world processes by extracting knowledge from modern information systems. Through data mining and machine learning techniques, PM reveals valuable insights into the processes of organizations, such as discovering process models, detecting inefficiencies, and predicting future behavior. 

Predictive Process Monitoring (PPM) is a subfield of PM that focuses on providing insights into ongoing business processes, including predicting the remaining time, the next event, or the outcome of a process. The latter field is also known as Outcome-Oriented Predictive Process Monitoring (OOPPM) and aims to predict the future state of ongoing cases. For instance, in a loan application process, all the events record specific activities-referred to as control flow attributes—leading up to the acceptance or rejection of the loan request. The primary goal of OOPPM is therefore to address critical questions, such as whether the loan application will be approved.  This means that it is required to monitor and analyze a vast amount of data from complex systems, such as financial transactions or patient admissions. In this context, it is essential for predictive models to effectively handle malicious, faulty, noisy, or unpredictable inputs. Adversarial testing has emerged as a valuable approach for assessing model robustness under such challenging data conditions \cite{DBLP:conf/bpm/VenkateswaranMI21, DBLP:conf/icml/SrivastavaHL20, DBLP:journals/jiis/PeeperkornBW23, stevens2023manifold, stevens2022assessing, DBLP:conf/nips/ChenZS0BH19}, addressing trustworthiness concerns in critical decision-making contexts \cite{stevens2022assessing, DBLP:conf/nips/ChenZS0BH19}.

\IEEEpubidadjcol %used to avoid overlap with the IEEE number

This work extends research \cite{stevens2022assessing, stevens2023manifold} on the vulnerability of OOPPM methods to input perturbations. In \cite{stevens2022assessing}, we already assessed how adversarial noise affects OOPPM predictive performance and explanation robustness across training and testing data. In this work, we introduced adversarial noise to the input data by perturbing dynamic event attributes in business process data, similar to subtle pixel changes in images that can significantly influence predictions. For example, in a loan application, modifying dynamic activity attributes like the \emph{Credit Score} and \emph{Income Level} affected the outcome predictions. The limitation of this work was that we avoided permuting the control flow attributes, as changes to activities and their position within a case are inherently noticeable due to the discrete nature of sequences. Even small modifications, such as substituting activities with those in nearby positions in the embedding space, such as moving \emph{Credit Check} before \emph{Document Verification}, can lead to unrealistic scenarios, as the sequence of events in a business process is often critical to its execution. In \cite{stevens2023manifold}, we addressed the prior limitation of avoiding perturbations to control flow attributes with the use of manifold learning, by projecting the generated adversarial examples to the learned class-specific data distributions, which means that the adversarial examples are more likely to preserve the essential characteristics of the data. However, this method does not guarantee that the adversarial examples are indistinguishable from the original business process cases, as multiple activity changes across various event positions may have happened.

\begin{figure}[ht]
    \centering    \includegraphics[width=\linewidth]{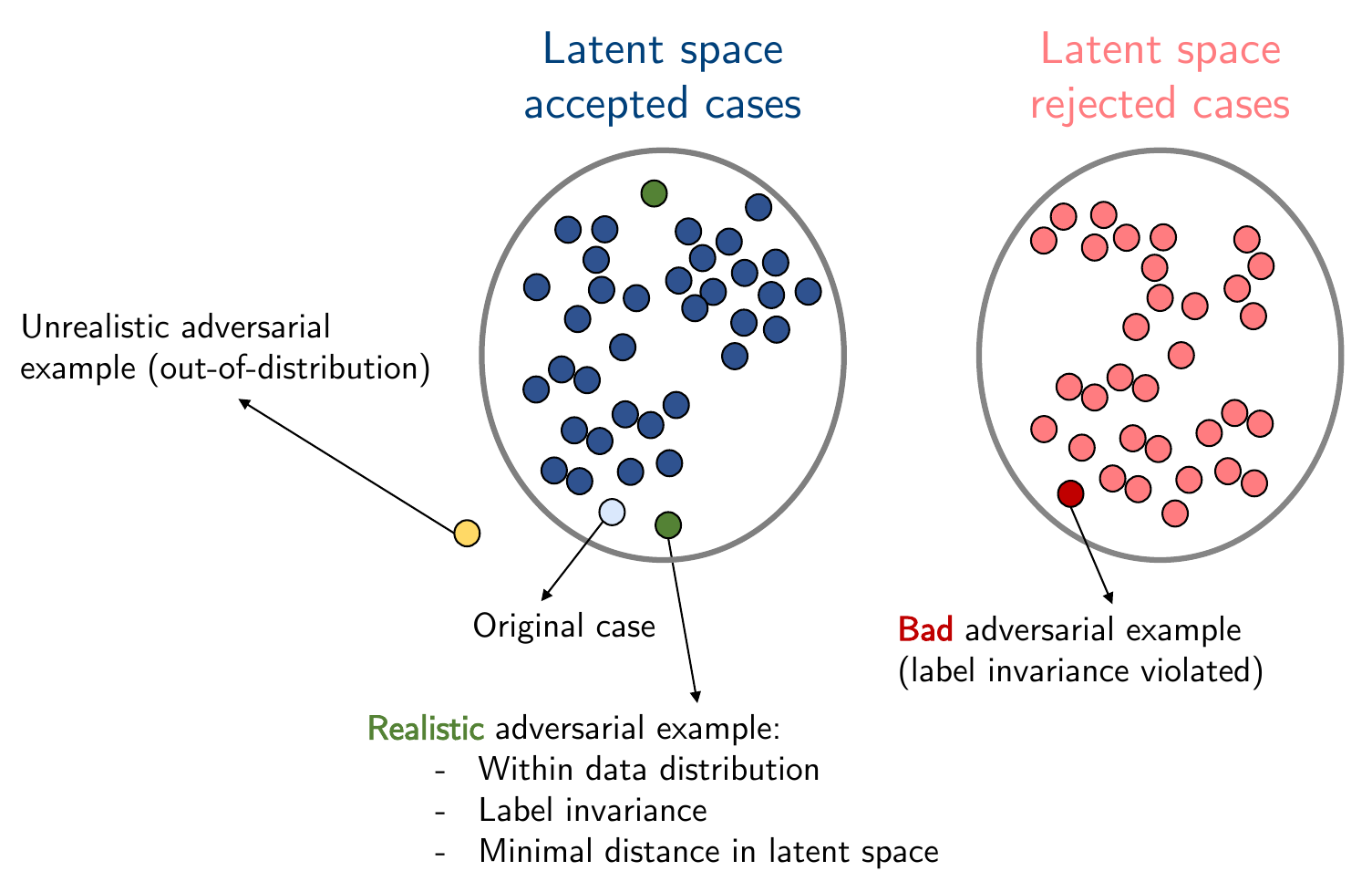}
    \caption{An illustrative example of adversarial examples in latent space. This simplified scenario assumes two perfectly separated data distributions.}
    \label{fig: advlatentspace}
\end{figure}

In this paper, we improve, extend, and benchmark these methods with novel adversarial attacking methods. 
The \emph{latent space attacks} (i.e. the \emph{latent sampling attack} and the \emph{gradient steps attack} identify adversaries within a dense and continuous representation space of the data rather than within the input data space. Latent space representations are able to capture high-level structural or sequential patterns within the data, and therefore preserve relationships between activities that would be lost in direct sequence manipulation. Furthermore, the generation of these latent space attacks is domain-agnostic and does not require specific process knowledge, as we introduce noise directly into the latent space representation of business processes, specifically targeting discrete event sequences in PPM. Unlike traditional methods that require direct manipulation of original event sequences, our latent space attacks generate adversarial examples from this compressed representation, producing realistic event sequences, while increasing the success rate of the attacks. Although previous studies have highlighted the effectiveness of perturbations within latent spaces~\cite{zhao2018generating,alzantot2018generating}, these methods are not capable of handling discrete event sequences. We address this gap by encoding business processes into a continuous latent space, and we generate adversarial examples within this space and map them back to the original input data space. In this way, we are able to manipulate event sequences without relying on domain-specific knowledge, a capability not covered by prior research. Figure \ref{fig: advlatentspace} illustrates how such adversarial examples look in latent space. The gradient steps approach, however, is a white-box attacking method and is used for benchmark purposes only. 

Next, we provide an extensive benchmark of eight different attacking methods, with multiple black-box and white-box adversarial attacks on business process executions, assessing model vulnerability through attack success rates. The successfulness of an attack is dependent on the difficulty of the adversarial attack, and the robustness of the model.

To summarize, we investigate the following research questions:
\begin{questions}
    \item How can we generate realistic adversarial examples in the latent space, while conforming to the same data distribution and real-world process constraints, without relying on domain-specific knowledge?
    \item How vulnerable are OOPPM models to these \emph{realistic} adversarial attacks? 
\end{questions}

The rest of this work is structured as follows. In section \ref{sec: background}, background information on adversarial machine learning (in PPM) is given. Next, section \ref{sec: methodology} provides information on the eight adversarial attack methods. section \ref{sec: results} reports on an experimental evaluation with eleven real-life event logs and four different classifiers. In section \ref{sec: relatedwork}, the related work is discussed. section \ref{sec: Robustconclusion} concludes the paper and points to elements for future work.

{\section{Background}\label{sec: background}

This section provides an overview of the preprocessing steps needed for OOPPM purposes and discusses key concepts from the field of adversarial machine learning and manifold learning.

\subsection{Predictive Process Monitoring}
An event log $L$ consists of events grouped by cases, forming process execution traces. An event $e$ from the event universe $\xi$ is represented as a tuple $e = (c,a,t,d,s)$, where $c \in C$ denotes the case ID, $a \in A$ represents the activity (i.e., the \emph{control flow} attribute), and $t \in \mathbb{R}$ indicates the timestamp. Each event also includes event-related attributes, or \emph{dynamic} attributes, which can vary during the course of the case and are represented by $d = (d_{1},d_{2},\dots,d_{m_{d}})$. Attributes that have the same value for every event with the same case identifier are called \emph{static} attributes, represented by $s = (s_{1},s_{2},\dots,s_{m_{s}})$. A trace is a sequence of events $\sigma_c = [e_{1},e_{2},\dots,e_{i},\dots,e_{n}]$, with $c$ as the case ID and $i$ as the index in the trace, ordered by event timestamps. An event $e_{i}$ in case $j$ of the event log $L$ is denoted as $e_{i,j} = (c_{j},a_{i,j},t_{i,j},d_{i,j}, s_{j})$. Each trace has a class label $y(\sigma_{c}) \in \mathcal{Y}$, with $\mathcal{Y}$ = \{0,1\} in the case of a binary outcome. This class labeling depends on the needs and objectives of the process owner \cite{DBLP:journals/tkdd/TeinemaaDRM19}. An example event log is provided in Table \ref{tab: exampleeventlog}, containing the data about Tom (Case A) and Sarah (Case B). The outcome is defined by whether the loan application is \emph{accepted} or \emph{rejected}.

\begin{table}[ht]
\centering
\caption{Example event log}
\label{tab: exampleeventlog}
\resizebox{\columnwidth}{!}{%
\begin{tabular}{|
>{\columncolor[HTML]{EFEFEF}}r 
>{\columncolor[HTML]{FFFFFF}}l 
>{\columncolor[HTML]{FFFFFF}}c 
>{\columncolor[HTML]{FFFFFF}}c 
>{\columncolor[HTML]{FFFFFF}}c 
>{\columncolor[HTML]{FFFFFF}}c 
>{\columncolor[HTML]{FFFFFF}}c |}
\hline
\multicolumn{1}{|l}{\cellcolor[HTML]{EFEFEF}\textit{Case ID}} &
  \cellcolor[HTML]{EFEFEF}\textit{Activity} &
  \cellcolor[HTML]{EFEFEF}\textit{Timestamp} &
  \cellcolor[HTML]{EFEFEF}\textit{Event Nr.} &
  \cellcolor[HTML]{EFEFEF}\textit{Resource} &
  \cellcolor[HTML]{EFEFEF}\textit{Loan Amount} &
  \cellcolor[HTML]{EFEFEF}\textit{Outcome} \\
A & Enter Loan Application & 23:00 & 1 & R5 & 250,000 & Rejected \\
A & Initial Review         & 23:25 & 2 & R2 & 250,000 & Rejected \\
A & Document Verification  & 23:35 & 3 & R2 & 250,000 & Rejected \\
A & Credit Check           & 23:56 & 4 & R2 & 250,000 & Rejected \\
B & Enter Loan Application & 23:00 & 1 & R1 & 390,000 & Accepted \\
B & Initial Review         & 23:10 & 2 & R1 & 390,000 & Accepted \\ \hline
\end{tabular}%
}
\end{table}

To progressively learn from different stages of traces, a prefix log $\mathbb{L}$ is extracted from the event log $L$, encompassing all prefixes from the complete traces $\sigma$. A trace prefix of case $c$ with length $l$ is defined as $\sigma_{c,l} = [e_{1},e_{2},\dots,e_{l}]$, where $l \leq |\sigma_c|$. Given the varying lengths of each (prefix) trace, a sequence encoding mechanism is necessary when using machine learning models with diverse attribute quantities. The aggregation encoding mechanism from \cite{de2016general} employs one-hot encoding to convert a dynamic categorical attribute into several transformed attributes. Frequency vectors are then extracted for each unique attribute value, representing the frequency of occurrence in the (prefix) trace. 

Step-based models such as recurrent neural networks can use the activity labels and dynamic attributes associated with subsequent events one by one, preserving the sequential nature of the process data, and allowing the model to effectively capture its inherent temporal relationships. Categorical features are often converted into a set of binary vectors through one-hot encoding.
To provide a low-dimensional representation of discrete attributes, embeddings are often used. This method transforms categorical attributes into a vector of continuous numbers meaningfully, avoiding the high-cardinality issue and nonsensical vector distances associated with one-hot encoding. The transformed attributes are denoted by $(x_{1},x_{2},\dots,x_{p})$. The transformed prefix log is used to train a predictive model $F$, which predicts the prefix (trace) $i$ as $\hat y_i = F(x_{i,1},\dots,x_{i,p})$. 

\subsection{Adversarial Machine Learning}
In the context of machine learning, addressing vulnerabilities against adversarial attacks can be an important tool for mitigating the risks posed by adversaries who exploit model weaknesses, or test the ability to reliably perform on possible out-of-domain samples or unseen in-domain samples that were not present during training.

This latter perspective on robustness ensures that machine learning models can generalize effectively across a variety of different inputs.

\begin{figure}
    \centering    \includegraphics[width=1\linewidth]{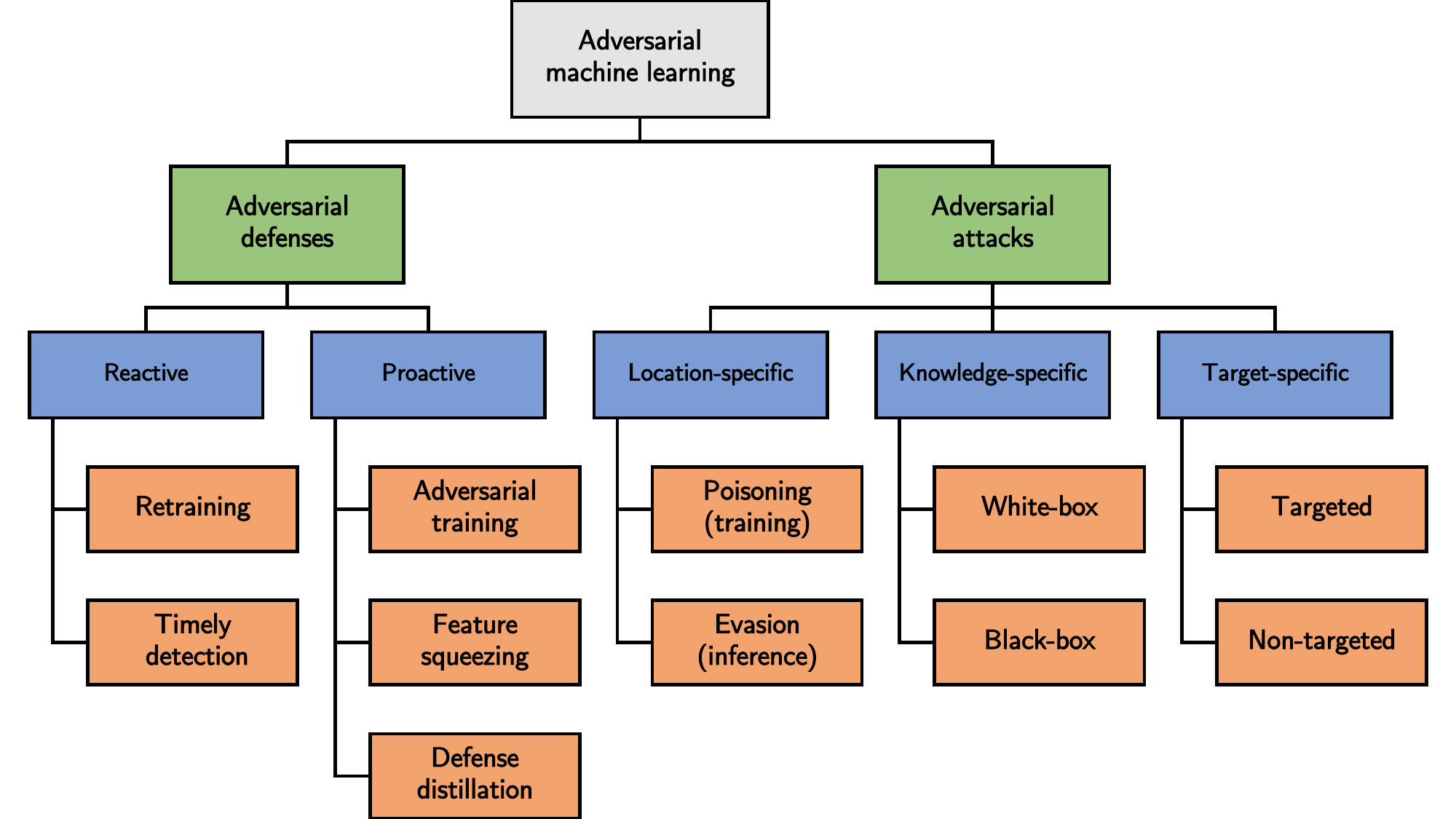}
    \caption{A taxonomy of adversarial attacks and adversarial defenses in AML. This figure is adapted from \cite{usama2019adversarial}.}
    \label{fig: taxonomyattacks}
\end{figure}

Adversarial Machine Learning (AML) focuses on examining the vulnerability against adversarial attacks, and their ability to withstand both adversarial perturbations and unexpected variations in data. In \cite{usama2019adversarial}, a taxonomy of adversarial defenses and attacks is introduced to the field of AML, providing a structured framework for categorizing various vulnerabilities and defense mechanisms. 
This taxonomy is visualized in Figure \ref{fig: taxonomyattacks}, showing both adversarial defenses and attacks. However, this paper focuses exclusively on adversarial attacks. The adversarial attacks are divided into three categories: \textit{location-specific}, \textit{knowledge-specific} and \textit{intent-specific}. Poisoning (evasion) attacks are when the training (testing) data is perturbed to mislead the ML model into incorrect predictions. 
In this paper, we only focus on evasion attacks. Next, adversarial attacks can be categorized into two knowledge-specific types: white-box attacks and black-box attacks. 

In white-box attacks, the attacker has full access to the model, including its architecture, parameters, and training data. This allows them to exploit this detailed information to craft highly targeted attacks. In contrast, black-box attacks are conducted with limited access, relying only on the output or predictions of the model without insight into its internal workings. In this paper, we employ both black-box methods (e.g., the \emph{regular attack} \cite{stevens2022assessing}, which use only the output of the model to guide the attack), and white-box methods (e.g., the \emph{gradient-based attack}, which leverage full access to the internal gradients, see Section \ref{sec: methodology}). Additionally, attacks can also be divided into \textit{targeted} and \textit{non-targeted attacks}. Non-targeted attacks, such as regular adversarial examples, are designed to induce misclassification without steering the prediction toward a specific class. On the other hand, targeted attacks, like manifold-based attacks \cite{stevens2023manifold}, aim to intentionally mislead the model into predicting a specific, often predefined, class or outcome. Note that, as we are in a binary classification setting, our non-targeted regular adversarial attack will result in the same outcome as it would in a targeted fashion. In this paper, we focus exclusively on adversarial attacks on the test data (evasion attacks) to ensure consistent and comparable results. By keeping the training dataset and model unchanged throughout the experiments, we eliminate variability introduced by attacking the training process (poisoning attacks). This approach allows for a direct comparison of performance across different attack strategies and methods on the same test dataset, ensuring that observed differences arise from the nature of the attacks rather than changes in the training configuration.

Two types of adversarial attacks have already been introduced to OOPPM \cite{stevens2022assessing}, the \emph{$last$ $event$ attack (A1)} and the $all$ $events$ attack (A2). In \cite{stevens2022assessing} the A1 attack targets the last event in the trace, modifying its dynamic attributes, resulting in a trace $\sigma_c^{A1} = [e_{1}, e_{2}, ..., e_{i}, ..., e_{n}^{A1}]$. 
In this paper, however, we focus exclusively on permuting the \emph{activity} values within each trace. This approach contrasts with \cite{stevens2023manifold}, where both the \emph{activity} and \emph{resource} attributes can be permuted. In this paper, the final event therefore looks like this: $e_{n}^{A1} = (c, a^{A1}, t)$. In \cite{stevens2022assessing} and \cite{stevens2023manifold}, the A2 attack modifies the dynamic attributes of all events in the trace, resulting in $\sigma_{c,l}^{A2} = [e_{1}^{A2}, e_{2}^{A2}, ..., e_{i}^{A2}, ..., e_{l}^{A2}]$, where $e_{i}^{A2} = (c_{j}, a_{j}, t_{j}, d_{j}^{A2})$. In this paper, we consider the A2 attack as: $\sigma_{c,l}^{A2} = [e_{1}^{A2}, e_{2}^{A2}, ..., e_{i}^{A2}, ..., e_{l}^{A2}]$, where $e_{i}^{A2} = (c_{j}, a_{j}^{A2}, t_{j})$

Valid adversarial attacks must adhere to the label invariance assumption, which states that the true class label of a data point remains unchanged even if the input data have been perturbed or manipulated. In other words, a theoretically perfectly robust and reliable model (or classifier) should not alter its predicted label in response to changes (attacks) made to the input.

\emph{Label invariance assumption.}
The label invariance assumption refers to the property that the underlying class label of a data point remains unchanged even if the input data have been perturbed or manipulated, such as in the case of adversarial attacks. 

The label invariance assumption can be expressed as:
\begin{equation}
y = y^{A*}, \quad f(\mathbf{\sigma_{c}}) \neq f(\mathbf{\sigma_{c}^{A*}}), \quad f(\mathbf{\sigma_{c}}) = y
\end{equation}

Consider an original trace $\sigma_c$ and its true label $y$. Let $\sigma_c^{A*}$ be an adversarial example generated from $\sigma_c$ by a certain adversarial attack. The true labels of both $\sigma_c$ and $\sigma_c^{A1}$ should be the same, but the predicted labels $f(\sigma_c)$ and $f(\sigma_c^{A*})$ could differ. Additionally, the original example $\sigma_c$ must be correctly predicted, i.e., $f(\mathbf{\sigma_c}) = y$. An adversarial attack is considered \emph{successful} when the original prediction made by the model changes.

For image data, the concept of adversarial examples is relatively straightforward. The visual features can be perturbed with imperceptible noise, leading to misclassifications while maintaining the original label. In contrast, generating adversarial examples for process data presents more significant challenges. Process data is structured, typically temporal, and context-dependent. Perturbations must carefully maintain the underlying control flow and domain-specific constraints of the process, making it difficult to ensure that the adversarial example remains a valid process instance. This complexity arises because the relationships between events and activities are critical, and perturbations can easily break these relationships, leading to implausible attacks, possibly not in line with the overall or the class-specific process behavior. 

\emph{Successful Adversarial Attack.}
An adversarial attack is successful when the original prediction made by the model has changed. The \textit{success rate} therefore measures how many adversarial examples have misled the model relative to the number of input data points. By contrast, a predictive model is considered \emph{robust} against minor input changes if the predicted labels of the original and adversarial instances remain the same. 

\emph{Adversarial robustness.}
Adversarial robustness is the ability of the model to maintain predictive performance despite being presented with adversarial examples designed to deceive it. Early theories attributed this vulnerability to rare cases \cite{DBLP:journals/corr/SzegedyZSBEGF13} or the linear nature of deep learning networks \cite{DBLP:journals/corr/GoodfellowSS14}. However, recent works state that regular adversarial examples are seen as unlikely in real-life scenarios and often lack semantic meaning \cite{zhao2018generating,schott2019towards}. This has led to a focus on generating more natural, \emph{on-manifold} adversarial examples, which are examples generated within the data manifold, making them more realistic and relevant to real-world scenarios. Some works even suggest that regular adversarial examples deviate from the manifold, and can therefore change their true label concerning the data distribution, potentially violating the label invariance assumption \cite{DBLP:conf/cvpr/Stutz0S19,DBLP:conf/iclr/GilmerMFSRWG18}.

\subsection{Manifold Learning}

The manifold hypothesis \cite{connor2021variational} posits that high-dimensional data can be effectively represented on a lower-dimensional manifold, an idea central to various methods aimed at ensuring adherence to the data distribution \cite{DBLP:joshi, DBLP:conf/www/pawelczyklearning, schott2019towards, DBLP:conf/nips/DhurandharCLTTS18,van2021interpretable, van2021conditional}. Generative models such as Variational Autoencoders (VAE) and Generative Adversarial Networks (GANs) are used in manifold learning to map low-dimensional latent vectors to high-dimensional data while adhering to an assumed prior distribution, typically Gaussian \cite{kingma2013auto}. We refer to this process as \emph{manifold learning}. 

Unlike a vanilla (non-variational) autoencoder that focuses on reconstructing the input \(x\) through the encoding process of an encoder \(e\) to generate a latent representation \(z = e(x)\), VAEs frame this as a probabilistic challenge. The objective of a VAE is to maximize the Evidence Lower Bound (ELBO), balancing the reconstruction loss (\emph{"how accurately can we reconstruct the original data point?"}) with the Kullback-Leibler (KL) divergence (\emph{"how closely does the learned latent variable distribution match the chosen prior distribution?"}). In VAE training, the prior distribution \(P\) is often the standard normal distribution, and the learned latent variable distribution \(Q\) is the distribution approximated by the VAE encoder.

The reconstruction loss is typically measured by the Negative Loss-Likelihood (NLL) loss:

\begin{equation}
\begin{split}
L(\theta)_{NLL} = -\sum_{\sigma_c \in L} & y_{\sigma_c} \cdot \log(\hat{y}_{\theta,\sigma_c}) \\
& + (1 - y_{\sigma_c}) \cdot \log(1 - \hat{y}_{\theta,\sigma_c})
\end{split}
\label{eq:NL}
\end{equation}

The notation $\hat{y}_{\theta,\sigma_c}$ is used to emphasize the dependency of the predicted probability on the model parameters $\theta$. The KL divergence is the sum taken over all components of the distributions (i.e. the number of latent variables in the latent space, $r$, and $r \leq p$),  where \(r \leq p\)):

\begin{center}
\begin{align}
L_{KL}(P || Q) = \sum_{\sigma_c \in L} P(\sigma_c) \cdot \log\left(\frac{P(\sigma_c)}{Q(\sigma_c)}\right) \label{eq:DKL}
\end{align}
\end{center}

In the context of OOPPM, class-specific manifolds are learned by creating manifolds specifically for the prefixes of each class label, as introduced in \cite{stevens2023manifold}. The Long Short-Term neural network (LSTM) VAE is thus trained using an LSTM-based architecture on these class-specific prefixes, which leverages the sequential information inherent in process data. By focusing on the prefix data for each outcome class separately, the model more effectively captures the specific temporal dynamics and uncovers the underlying structure or manifolds of high-dimensional data specific to certain classes, while discarding noise and redundancy. Within this setup, the encoder networks compute a low-dimensional latent representation of data, and the decoder network reproduces samples from the latent space. By imposing a prior over the latent distribution $p(z)$, the variational aspect of the VAE regularizes the encoder and structures the latent space. 

\section{Generating Adversarial Examples for Outcome-Oriented Predictive Process Monitoring}\label{sec: methodology}
\begin{figure}
\centering
\scalebox{0.7}{
\begin{minipage}{1.40\columnwidth}
\begin{algorithm}[H]
\caption{The algorithm to generate class-specific adversarial examples based on the attack strategy, attack type, number of adversarial examples and number of events (optional). Note that we disregard the initialization of variables for simplicity reasons.}
\label{alg: adversarial_examples}
\begin{algorithmic}[1]
\Require $\mathit{prefixes}$, $\mathit{type_{attack}}$, $\mathit{strategy_{attack}}$, $\mathit{nr_{adv}}$, $\mathit{k_{events}}$
\State
\State \Comment{\textbf{For each prefix, we generate adversarial examples}}
\For{$\mathit{prefix} \in \mathit{prefixes}$}
    \If{$\mathit{strategy_{attack}} == \mathit{latent\_sampled}$} 
        \State $\mathit{adv_{prefixes}} \gets \text{latent\_sampled}(\mathit{prefix}, \mathit{nr_{adv}}, \mathit{adv_{total}})$ 
    \ElsIf{$\mathit{strategy_{attack}} == \mathit{gradient\_based}$}
        \State $\mathit{adv_{prefixes}} \gets \text{gradient\_based}(\mathit{prefix}, \mathit{nr_{adv}}, \mathit{adv_{total}})$     
    \Else
        \If{$\mathit{type_{attack}} == \mathit{last\_event}$}
            \State $\mathit{adv_{prefixes}} \gets \text{permute\_last\_event}(\mathit{prefix})$ 
        \ElsIf{$\mathit{type_{attack}} == \mathit{all\_event}$}
            \State $\mathit{adv_{prefixes}} \gets \text{permute\_all\_event}(\mathit{prefix}, \mathit{nr_{adv}})$ 
        \ElsIf{$\mathit{type_{attack}} == \mathit{k\_event}$}
            \State $\mathit{adv_{prefixes}} \gets \text{permute\_k\_events}(\mathit{prefix}, \mathit{k_{events}}, \mathit{nr_{adv}})$ 
        \EndIf
        \EndIf
        \If{$\mathit{strategy_{attack}} == \mathit{projected}$}
            \State $\mathit{mu_{adv}} \gets \text{encode\_on\_latent\_space}(\mathit{adv_{prefixes}})$ 
            \State $\mathit{adv\_prefixes} \gets \text{decode\_latent\_space}(\mathit{mu_{adv}})$
        
    \EndIf
    \State
    \State \Comment{\textbf{Find adversarial example with minimal distance in latent space}}
    \State $\mathit{prefixes} \gets [\mathit{prefix}] + \mathit{adv_{prefixes}}$ 
    \State $\mathit{mu}, \_ \gets \text{encode\_on\_latent\_space}(\mathit{prefixes})$ 
    \State $\mathit{dist_{pair}} \gets [\|\mathit{mu}[0] - \mathit{sample}\| \text{ for sample in } \mathit{mu}[1:]]$ 
    \State $\mathit{adv_{prefix}} \gets \mathit{adv_{prefixes}}[\mathit{min_{index}}(\mathit{dist_{pair}})]$ 
    \State $\mathit{adv_{total}.append}(\mathit{adv_{prefix}})$ 
    \State $\mathit{distances_{total}.append}(\mathit{min_{index}}(\mathit{dist_{pair}}))$ 
\EndFor

\State \Return $\mathit{adv_{total}}$, $\mathit{distances_{total}}$ 
\end{algorithmic}
\end{algorithm}
\label{alg: adv_pseudocodeRobustPPM}
\vspace{-10mm}
\end{minipage}
}
\end{figure}
\begin{figure*}[ht]
    \centering    \includegraphics[width=0.8\linewidth]{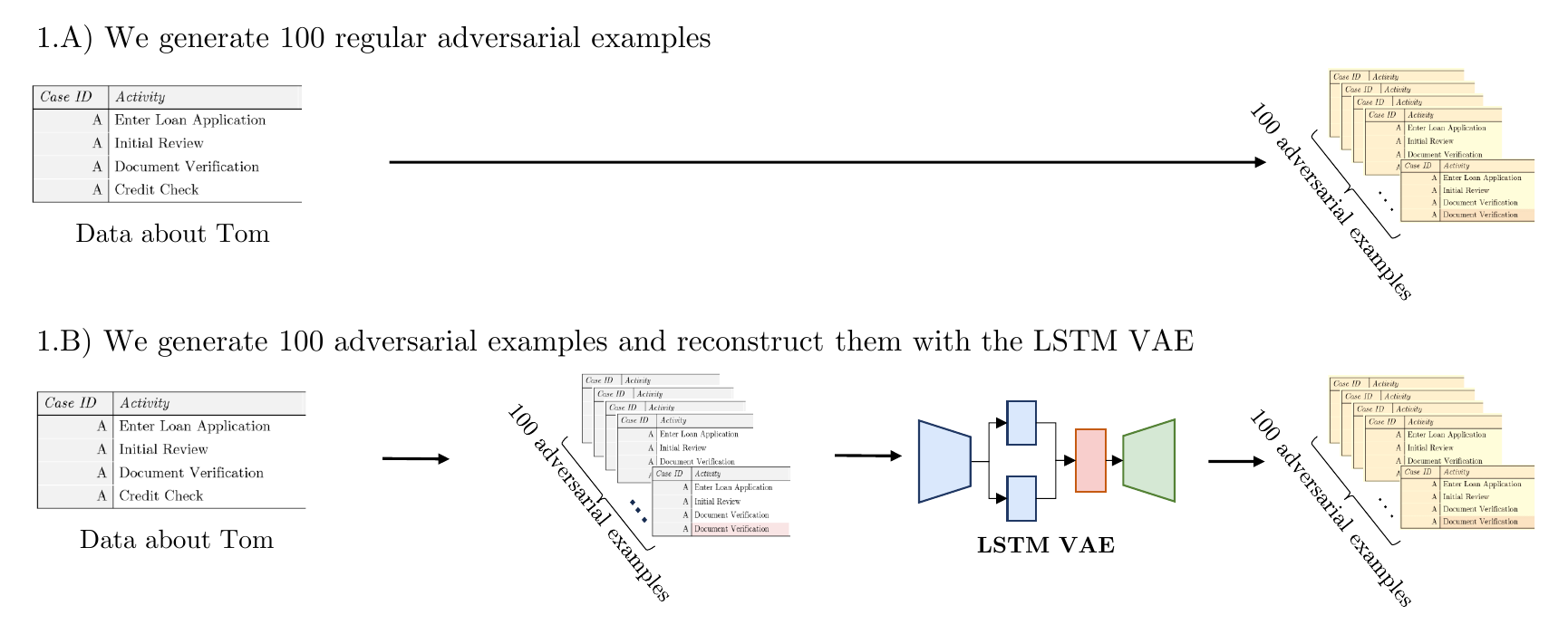}
    \caption{Illustration of (1.A) the regular adversarial attack introduced by \cite{stevens2022assessing}, and (1.B) the manifold-projected adversarial attack as described by \cite{stevens2023manifold}.} \label{fig: regularprojected}
\end{figure*}

In this section, we detail four attack strategies---\emph{regular}, \emph{projected}, \emph{latent sampling}, and \emph{gradient steps}---that together yield eight distinct adversarial attack methods. We explain how each attack strategy addresses the challenges posed by the sequential nature of process data. The \emph{regular} and \emph{projected} strategies operate directly within the original input space, representing \emph{input space attacks}. These two strategies employ three distinct approaches for attacking individual events within the data sequence, resulting in six methods. In contrast, the \emph{latent sampling} and \emph{gradient steps} strategies generate adversarial examples within the latent space, categorized as \emph{latent space attacks}. Together, these approaches result in eight unique adversarial attack methods.

\subsection{Adversarial Example Generation}\label{sec: adversarialexamplegeneration}

Algorithm \ref{alg: adversarial_examples} presents a general pipeline for generating adversarial examples, applicable across multiple attack strategies. This process begins with an iteration over prefixes (line 3), followed by the application of specific attack techniques (lines 4-20), which is further described in the subsequent subsections. Next, we select the adversarial prefix that is closest to the original in latent space, using pairwise Euclidean distances to measure the proximity between the latent representation of the original trace and the latent representation of the generated adversarial prefix (lines 22-29).

\paragraph*{Input Space Attacks}

The \emph{regular attack} strategy, illustrated in Figure \ref{fig: regularprojected}, follows the approach introduced in \cite{stevens2022assessing}. In this strategy, we directly attack the original trace, where the resulting adversarial example is the modified trace after applying the attack. We distinguish between three types of regular attacks: the \emph{last event attack (A1)}, which modifies only the last event of the trace (introduced in \cite{stevens2022assessing}); the \emph{all event attack (A2)}, which modifies every event (introduced in \cite{stevens2022assessing}); and the novel \emph{k event attack (A3)} targets a selected subset of $k$ events within the trace. 
The \emph{k event attack} is similar to the A1 and A2 attacks but additionally considers the positional constraints of activities within a case, making the attack more context-aware and reducing unrealistic perturbations that might otherwise break the realistic flow of the process. Specifically, this attack accounts for the fact that certain activities can only be executed (and therefore recorded as events) at specific positions within a process execution, performing perturbations only when an activity has already occurred in a similar position within the case. The A3 attack therefore only modifies \emph{activity} values based on tuples of (index, activity) extracted from the event log. This allows us to randomly select $k$ tuples, where each selected \emph{activity} of the tuple being permuted if (1) the \emph{activity} differs from the original and (2) the index has not already been modified, ensuring relevance to frequently occurring activities and contextual appropriateness.

Formally, given a trace $\sigma_c = [e_{1}, e_{2}, \ldots, e_{i}, \ldots, e_{n}]$, the A3 attack produces a modified trace:

\[
\sigma_c^{A3} = [e_{1}^{A3}, e_{2}^{A3}, \ldots, e_{i}^{A3}, \ldots, e_{n}^{A3}],
\]

where for each modified event \( e_{i}^{A3} \), we have:

\[
e_{i}^{A3} = 
\begin{cases} 
      (c_i, a_i^{A3}, t_i) & \text{if } e_i \text{ is selected for modification} \\
      (c_i, a_i, t_i) & \text{otherwise}.
\end{cases}
\]

In contrast, the \emph{projected attack} strategy, also shown in Figure \ref{fig: regularprojected}, builds on the concept of on-manifold attacks introduced in \cite{stevens2023manifold}. In this approach, we learn class-specific encoders and decoders that approximate the data distribution for each class by training them on class-specific prefixes. Then, after attacking the original trace, we project the adversarial example onto the class-specific data distribution. This approach adheres to the label invariance assumption, as the generated adversarial examples for a particular class remain within the boundaries of the data distribution of that class. This is in contrast to the traditional attacks that might result in unrealistic adversarial examples that would never occur in practice. 
An important consequence of this projection is that the length of the adversarial prefix may differ from the original prefix length, i.e. projecting an adversarial example onto this manifold can result in either longer or shorter adversarial examples after decoding. This is due to our approach of masking out the padding tensor (see section \ref{subsec: implementation}). What we could have done to prevent this is, adding a constraint for preserving a similar prefix length relative to the original trace. Another option could be to enforce sparsity in the generated examples. Enforcing sparsity would mean explicitly encouraging the generated examples to differ minimally from the original trace, typically by modifying only a small number of attributes to preserve as much of the original sequence as possible. Nonetheless, our primary objective in this paper is to generate successful and realistic adversarial examples rather than to produce adversarial examples of identical lengths.

\paragraph*{Latent Space Attacks} Different works argue that adding noise in the latent space allows for the generation of more natural adversarial examples, compared to adding noise in the input space. Many works argue that rather than permuting unimportant background pixels (in the case of image-based adversarial attacks), adversarial examples should be generated through perturbations in the latent space \cite{hendrycks2021natural, schott2019towards}. The first novel attacking method we propose is the \emph{latent sampling attack}, displayed in Figure \ref{fig: latentgradient}, and it is a black-box method that exploits the stochasticity introduced by the variational inference of the LSTM VAE manifold to create adversarial examples. This technique ensures that the instance stems from the same mean and variance as the original instance. Specifically, the reparameterization trick generates $\mathbf{z} = \boldsymbol{\mu} + \epsilon \cdot \boldsymbol{\sigma}$, where $\mathbf{z}$ is the latent variable, $\boldsymbol{\mu}$ is the mean, $\boldsymbol{\sigma}$ is the standard deviation, and $\epsilon$ is a random variable sampled from a standard normal distribution. This approach ensures that the adversarial example is generated from the same latent space representation, making the generated adversarial examples more \textit{realistic}.
\begin{figure*}[hb]
    \centering    \includegraphics[width=0.87\linewidth]{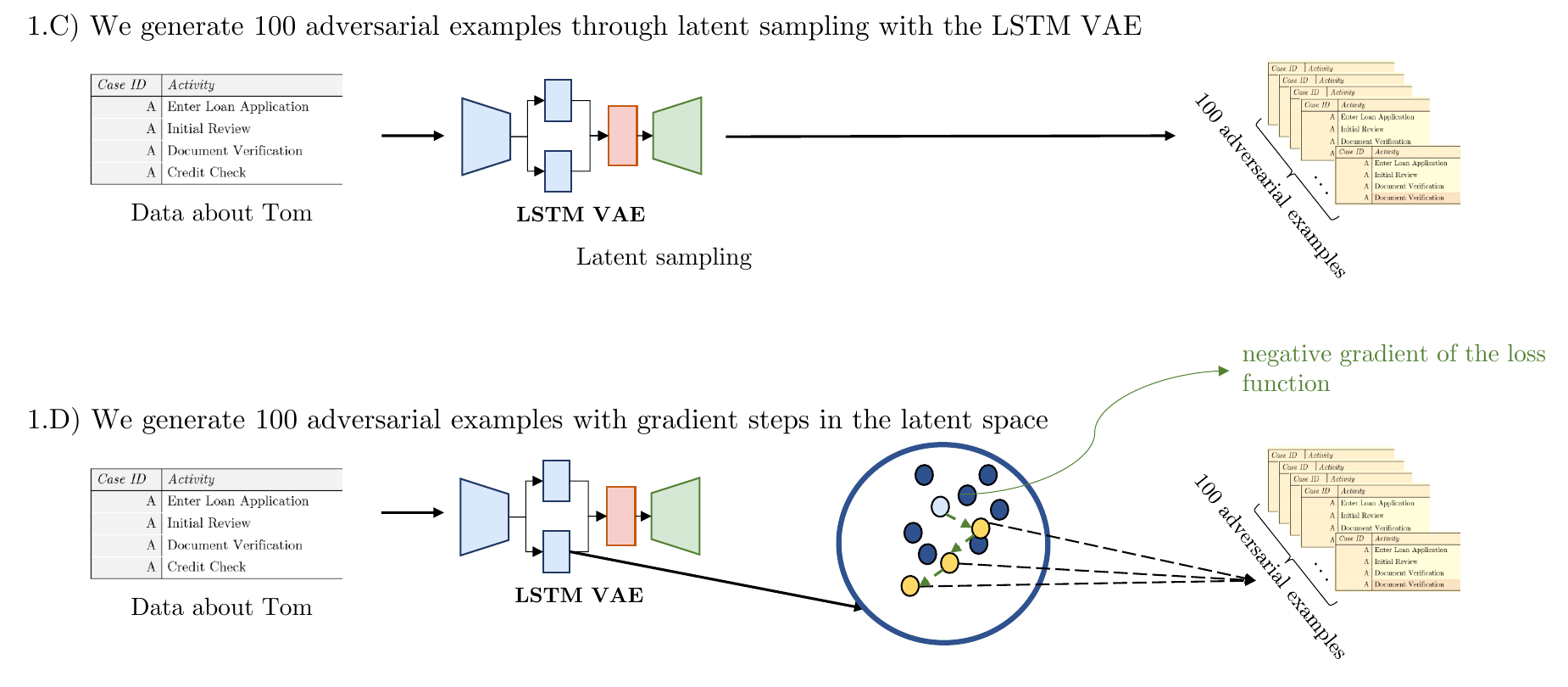}
    \caption{Illustration of (1.C) the latent sampling attack, which uses the stochasticity of the reparameterization trick to generate adversarial examples, and (1.D) the newly introduced gradient steps attacking strategy, which uses the gradients of the classifier to walk in the latent space, across the decision boundary.}
    \label{fig: latentgradient}
\end{figure*}

Conversely, the \emph{gradient steps attack}, displayed in Figure \ref{fig: latentgradient}, is a white-box approach that utilizes the gradients of the classifier to iteratively move across the decision boundary. In each iteration, the algorithm calculates the gradients of the classifier with respect to the input trace, and it identifies the most impactful directions in the latent space to apply perturbations, creating adversarial examples that are likely to deceive the model while remaining close to the original trace in latent space. In this paper, we adapt the REVISE counterfactual generation method \cite{DBLP:joshi} to handle the sequential structure of business process executions. The adapted algorithm is tailored to create latent space representations based on the one-hot encoded structure of activity sequences, and the gradient-based optimization minimizes Euclidean distance in the latent space. This ensures adversarial examples with minimal perturbations by default. However, despite its effectiveness in creating adversarial examples, the gradient steps attack is primarily useful for benchmarking purposes and not for practical use in real-world settings. The key reason for this limitation lies in the white-box nature of the attack: it requires full access to the model, including the gradients of the classifier. In real-world scenarios, adversaries typically do not have this level of insight into the model, and thus cannot use such an attack strategy.
Similarly to the projected attack, the length of the adversarial prefix for latent space attacks can again differ from the original prefix length.

\subsection{Closest Adversarial Example Selection}\label{sec: minimization}

In VAEs,  the latent space is generally modeled as a multivariate Gaussian distribution. The encoder learns the mean and variance of the distribution, allowing the latent points to be smoothly distributed within a Euclidean space, often viewed as a continuous low-dimensional manifold \cite{chen2020learning}. This design enables the encoder to map inputs to specific points in the latent space, where the Euclidean distance between points reflects the similarity between their latent representations. Figure \ref{fig: minimizationlatent} provides an illustration of the motivation for minimizing distances in the latent space, while Figure \ref{fig: minimization} demonstrates the process of selecting an adversarial example based on the smallest distance in the latent space.

\subsection{Class Label Prediction of Adversarial Examples}

We then use a classifier to predict the label outcome of the adversarial examples. An attack is considered successful if the predicted label of the adversarial example differs from the original label of the prefix trace.

\begin{figure}[ht]
    \centering    \includegraphics[width=\linewidth]{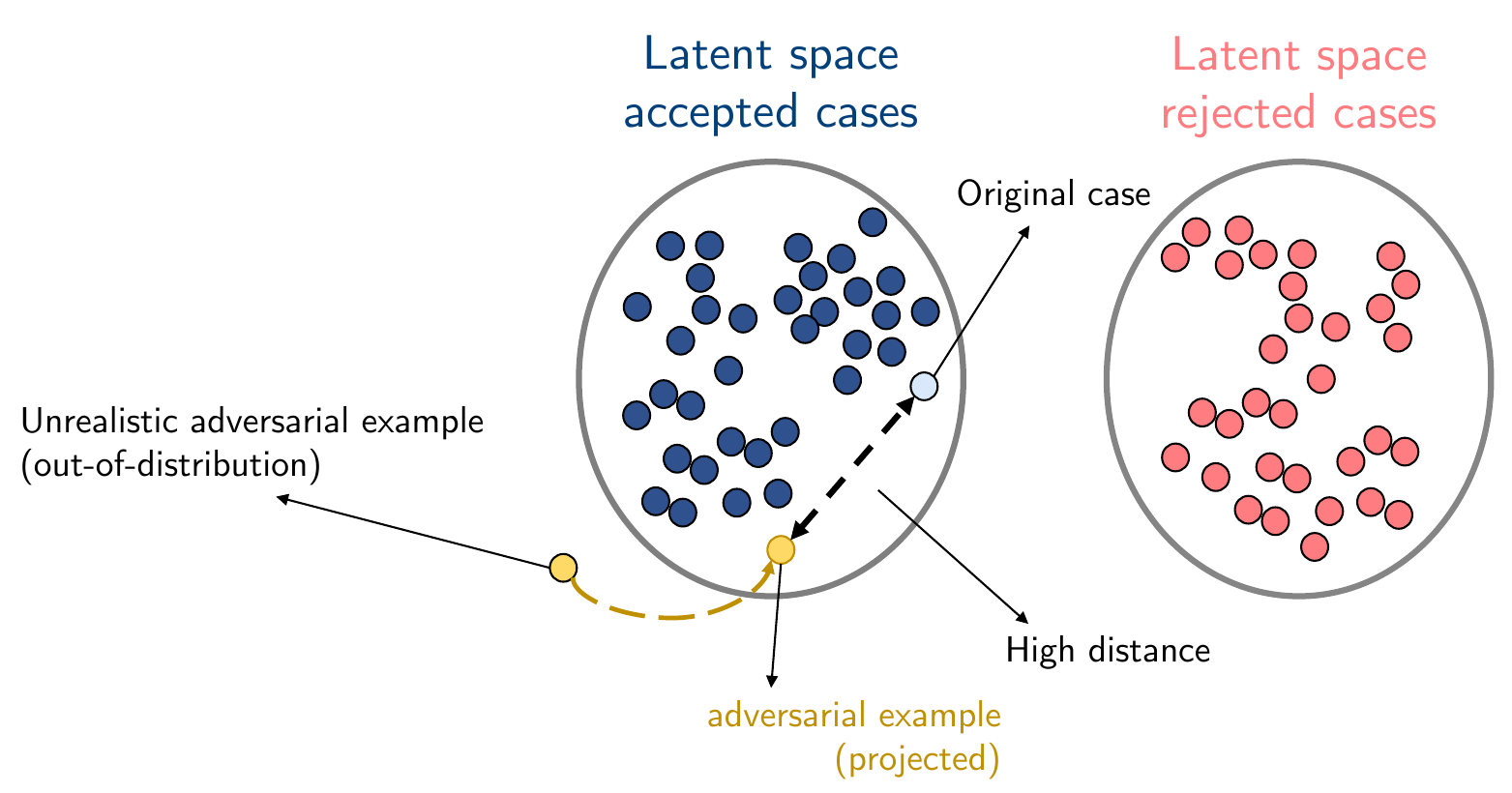}
    \caption{An illustrative example of adversarial examples in latent space. Note that projecting the regular adversarial example to the distribution does not ensure a minimal distance to the original example (in the latent space).}
    \label{fig: minimizationlatent}
\end{figure}

\begin{figure*}[ht]
    \centering    \includegraphics[width=0.8\linewidth]{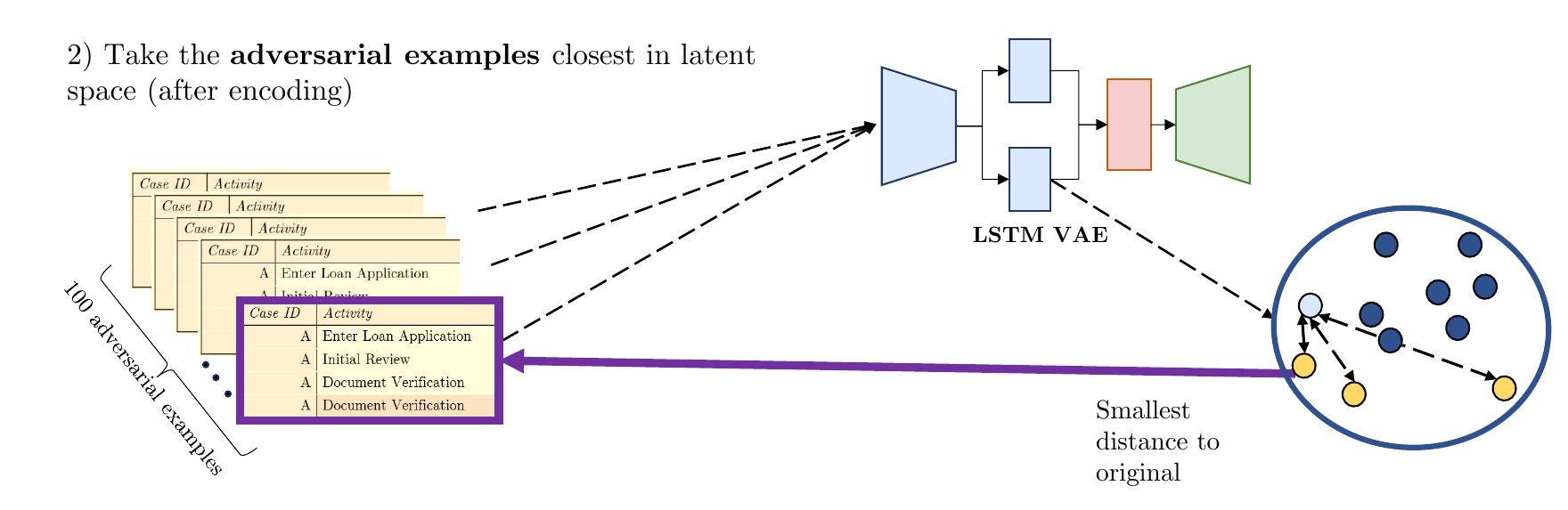}
    \caption{Illustration of (2) which is the minimization in the latent space as an additional requirement to adhere to the label invariance assumption.}
    \label{fig: minimization}
\end{figure*}

\section{Experimental Setup}\label{sec: results}

In this section, we describe the different event logs and their specifications, the benchmark models commonly used for OOPPM purposes, and the hyperoptimization settings and implementation details of various setups. 

\subsection{Event Logs}
\begin{table*}[ht]
\centering
\caption{The Different Specifications of the Event Logs.}
\label{tab:eventlogspecs}
\resizebox{0.9\textwidth}{!}{%
\begin{tabular}{|c|ccccccccccc|}
\hline
\rowcolor[HTML]{EFEFEF} 
Event Log &
  \textit{Traces} &
  \textit{Events} &
  \textit{Med.} &
  Min. &
  \textit{Max.} &
  \textit{Prefix} &
  \textit{Var.} &
  \textit{$\frac{Act}{Trace}$} &
  \textit{$\frac{Var.}{Trace}$} &
  \textit{$\frac{Events}{Trace}$} &
  \textit{Pos. Class Ratio} \\ \hline
\cellcolor[HTML]{C0C0C0}\textbf{BPIC2012 (1)} & 4685 & 67480  & 35 & 15 & 175 & 40 & 3578 & 36  & 0.76 & 14 & 0.48  \\
\cellcolor[HTML]{C0C0C0}\textbf{BPIC2012 (2)} & 4685 & 705476 & 35 & 15 & 175 & 40 & 3578 & 36  & 0.76 & 32 & 0.17  \\
\cellcolor[HTML]{C0C0C0}\textbf{BPIC2012 (3)} & 4685 & 70546  & 35 & 15 & 175 & 40 & 3578 & 36  & 0.76 & 15 & 0.35  \\
\cellcolor[HTML]{C0C0C0}\textbf{BPIC2015 (1)} & 696  & 28775  & 42 & 2  & 101 & 40 & 677  & 380 & 0.97 & 41 & 22.79 \\
\cellcolor[HTML]{C0C0C0}\textbf{BPIC2015 (2)} & 753  & 41202  & 55 & 1  & 132 & 40 & 752  & 396 & 1.00 & 55 & 61.29 \\
\cellcolor[HTML]{C0C0C0}\textbf{BPIC2015 (3)} & 1328 & 57488  & 42 & 3  & 124 & 40 & 1280 & 380 & 0.96 & 43 & 23.78 \\
\cellcolor[HTML]{C0C0C0}\textbf{BPIC2015 (4)} & 577  & 24234  & 42 & 1  & 82  & 40 & 576  & 319 & 1.00 & 42 & 38.56 \\
\cellcolor[HTML]{C0C0C0}\textbf{BPIC2015 (5)} & 1051 & 54562  & 50 & 5  & 134 & 40 & 1048 & 376 & 1.00 & 52 & 52.5  \\
\cellcolor[HTML]{C0C0C0}\textbf{SEPSIS (1)}   & 782  & 13120  & 14 & 5  & 185 & 29 & 684  & 14  & 0.87 & 16 & 0.54  \\
\cellcolor[HTML]{C0C0C0}\textbf{SEPSIS (2)}   & 782  & 10924  & 13 & 4  & 60  & 13 & 656  & 15  & 0.84 & 14 & 0.19  \\
\cellcolor[HTML]{C0C0C0}\textbf{SEPSIS (3)}   & 782  & 12463  & 13 & 4  & 185 & 22 & 709  & 15  & 0.91 & 16 & 0.2   \\ \hline
\end{tabular}%
}
\end{table*}

This research utilizes four distinct real-life event logs available at the 4TU Centre for Research Data website\footnote{https://data.4tu.nl/}, which are frequently employed in the domain of Outcome-Oriented Process Prediction and Monitoring (OOPPM) \cite{DBLP:journals/tkdd/TeinemaaDRM19, DBLP:journals/bise/KratschMRS21, DBLP:conf/icws/WangYLS19, DBLP:journals/jds/HarlWSM20, mehdiyev2021explainable}. The event logs are segmented using Linear Temporal Logic (LTL) rules as detailed in\cite{DBLP:journals/tkdd/TeinemaaDRM19}, establishing objectives for the processes. The segmentation is based on the labelling functions defined by the four LTL rules, resulting in four separate binary prediction tasks. Table~\ref{tab:eventlogspecs} provides the specifications of these event logs.

The first event log, BPIC2012, captures the execution of loan application processes in a Dutch financial institution. The event log includes the events related to a particular loan application, with the labeling indicating whether the final outcome of a case was either accepted, declined, or cancelled. Similar preprocessing steps for trace prefixing and cutting are employed as in~\cite{DBLP:journals/tkdd/TeinemaaDRM19}.

The BPIC2015 event log consists of events from the building permit application processes in five Dutch municipalities. An LTL rule is applied to the event log, with a split for each municipality. The rule stipulates that the activity \emph{send confirmation receipt} must always be followed by \emph{retrieve missing data}, and the latter must always occur if the former does. No trace cutting is performed on this event log.

The Sepsis cases event log contains discharge information for patients with sepsis symptoms in a Dutch hospital, from emergency room admission to discharge. Here, the labelling is based on patient discharge rather than LTL rules~\cite{DBLP:journals/tkdd/TeinemaaDRM19}.

\subsection{Benchmark Models}

Four different predictive models, which are commonly used in OOPPM \cite{DBLP:journals/tkdd/TeinemaaDRM19, DBLP:journals/bise/KratschMRS21}, e.g. Logistic Regression (LR), XGBoost (XGB), Random Forest (RF), and an LSTM neural network, are used to benchmark the findings. These models serve as the most representative of the statistical, machine learning, and deep learning facilities. LR is a transparent predictive model that is commonly used for classification, and both XGB and RF are popular tree-based ensemble models due to their fast convergence and high predictive accuracy. Finally, a long short-term memory LSTM neural network is used due to its popularity in OOPPM research \cite{weytjens2020process, kratsch2021machine}.

\subsection{Implementation}\label{subsec: implementation}

In this section, we describe the implementation details and assumptions made in this work. 

\subsubsection{Train-test Split}\label{sec: traintestsplit}
A temporal split is used to divide the event log into training and testing cases \cite{DBLP:journals/tkdd/TeinemaaDRM19}. The cases are ordered by start time, with the first 80\% used for training the predictive model and the remaining 20\% for evaluating its performance. This split is done at the level of completed traces, ensuring all prefixes of the same trace stay in the same set (either train or test). To avoid overlap, events in the training cases that occur during the test period are discarded. 

\subsubsection{Preprocessing Steps}

Then, we extract prefix logs and do the necessary preprocessing steps as detailed in \ref{sec: background}. For the machine learning models, aggregation encoding \cite{DBLP:journals/tkdd/TeinemaaDRM19} is used to encode the data for the machine learning algorithms. Note that aggregation encoding is unique to process data. On the other hand, the deep learning models are built to work with sequential models and therefore do not need the use of such a sequence encoding mechanism. 

\subsubsection{Model Training and Optimization}

There are two different types of models that need to be trained: a classifier and class-specific LSTM VAEs. The classifier is used to predict the outcome of the instances, whereas the LSTM VAEs are used to project or generate the adversarial examples.

The classifier is trained on the extracted prefix loss, The LSTM VAE, on the other hand, needs additional steps. First, to ensure that the LSTM VAE could learn the ending points of traces, we added an artificial End Of Sequence (EoS) token to all the (prefix) traces. During decoding, whenever the LSTM VAE produced the first EoS token, we masked the subsequent tokens with padding values. To prevent the model from learning irrelevant information from the length of the traces or the padding tokens, we masked the padding token value during backpropagation. It is important to note that this masking and the use of the EoS token were not necessarily for training and fitting the predictive models. Additionally, we provided the option to remove duplicate traces in the event logs, as duplicates are prevalent when only considering control flow. Furthermore, we allowed for the removal of ambiguous duplicates—identical traces with different labels, since such cases could mean that a trace serves as its own adversarial example. The implementation of the LSTM VAE is based on the work done in \cite{stevens2024generating}.

\subsubsection{Reproducibility of the Results}

The implementation of the benchmark setting is provided in the \href{https://github.com/AlexanderPaulStevens/AdversarialBP}{GitHub repository}, as we have transformed the extensive counterfactual generation benchmark named CARLA \cite{DBLP:conf/nips/PawelczykBHRK21}, for the use of adversarial example generation.
\section{Experimental Evaluation}\label{sec: evaluation}

In this section, we evaluate our work based on the answers provided to the research questions. 

The first research question \textbf{RQ1} can be answered with the use of section \ref{sec: methodology}, by showing how the two novel methods, i.e. the \emph{latent sampling attack} and the \emph{gradient steps attack} ensure that the generated adversarial examples conform to the original data distribution while adhering to real-world process constraints. Importantly, the proposed approaches circumvent the need for domain-specific knowledge by leveraging latent representations derived from generalizable process mining models.

The second research question \textbf{RQ2} evaluates and compares the performance of eight distinct attack strategies—both black-box and white-box—using metrics such as attack success rates and their implications on OOPPM methods. To address this, we calculate the success rate, as this is one of the most commonly used evaluation metrics to evaluate the vulnerability against adversarial examples \cite{xiao2018generating}. The \emph{success rate} on attacks is calculated by counting the percentage of adversarial examples that have a flipped prediction. A prediction is successfully flipped if the probability decreases/increases until it is lower/higher than the predefined optimal threshold for that classifier. Figure \ref{fig: successrateperlength} presents box plots of the success rates for the classifier and attacking method combinations, aggregated over the different event logs. Note that the gradient steps attack method was only applicable to the LSTM neural network. Even though XGBoost uses gradients as well, it is not inherently differentiable. Decision trees, and by extension XGBoost, work by recursively partitioning the input space and making piecewise constant decisions based on learned thresholds for the attributes. Since decision trees are piecewise constant, there is no continuous gradient to compute for the model with respect to the input data. The decision boundary of the model is therefore not smooth but is highly non-linear and discrete.

\begin{figure*}[ht]
    \centering
    \includegraphics[width=0.9\linewidth]{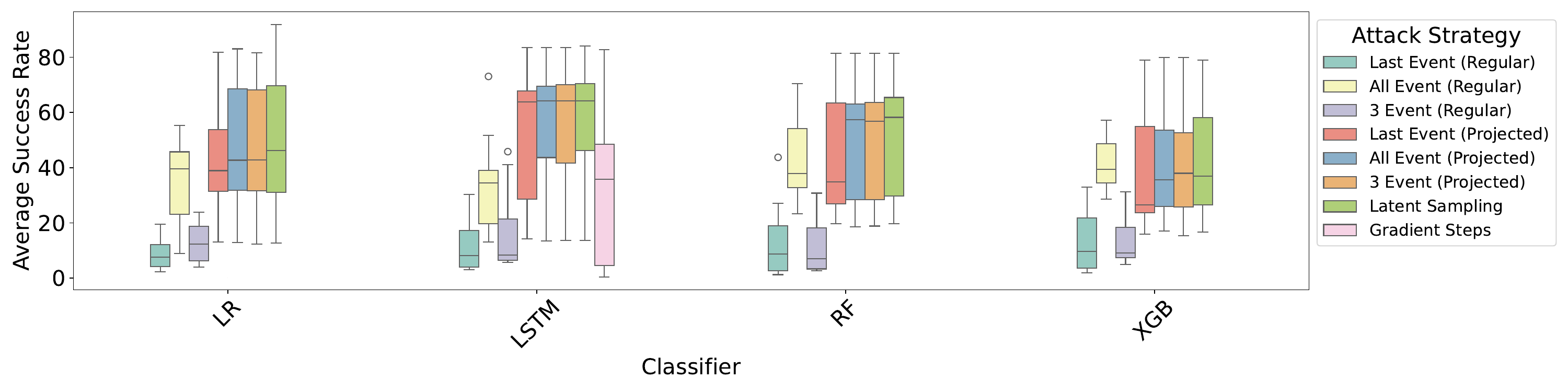}
    \caption{Average success rate for the different attack strategy combinations and the different classifiers, averaged over all the 11 event logs.}
    \label{fig: average success}
\end{figure*}

\subsection{Benchmark Results}

First, the classifiers can be ranked by their average AUC scores as follows: \emph{LR} with 80.24, \emph{LSTM} with 82.35, \emph{RF} with 83.93, and \emph{XGB} with 84.54. In contradiction with the findings of \cite{DBLP:journals/bise/KratschMRS21}, the DL model (LSTM) does not outperform the classical ML ensemble methods. First, we attribute the slight underperformance of the LSTM to the results for the event log \emph{BPIC2015 (2)} and \emph{SEPSIS (1)}. Note that the results for the SEPSIS logs are volatile in general. Second, the performance of the DL is not fully exploited due to the trace cutting, as the DL models are better capable of handling long-term dependencies.

The attack methods, ranked from least to most successful, are as follows:

(1) Last Event (Regular) with a success rate of 12.17\%

(2) 3 Event (Regular) with a success rate of 13.83 \%

(3) Gradient Steps with a success rate of 33.17 \%

(4) All Event (Regular) with a success rate of 38.05 \%

(5) Last Event (Projected) with a success rate of 45 \%

(6) All Event (Projected) with a success rate of 49.04 \%

(7) 3 Event (Projected) with a success rate of 50.08 \%

(8) Latent Sampling with a success rate of 50.17 \%.\\

The first insight from the ranking is that the \emph{Latent Sampling} method achieves the highest success rate. Additionally, the projected attack method introduced in \cite{stevens2023manifold} outperforms the regular attack method that was introduced in \cite{stevens2022assessing}. Increasing the number of iterations beyond the default 1500 could potentially further enhance success rates. Second, based on Figure \ref{fig: successrateperlength}, the LSTM seems to be the most vulnerable OOPPM model, as it consistently exhibits the highest average success rates across most attack methods. Furthermore, the \emph{Latent Sampling} attack method remains the most effective across all the four OOPPM methods. By contrast, the three regular attacking methods---\emph{Last Event (Regular)}, \emph{All Event (Regular)}, and \emph{3 Event (Regular)}---have the lowest success rate. Finally, Figure \ref{fig: average success} reveals an interesting pattern: for most adversarial example generation methods (excluding \emph{Last Event (Regular)} and \emph{3 Event (Regular)}), the number of successful adversarial examples increases with prefix lengths up to 30. This finding is counterintuitive, as one might expect that permutations in shorter prefixes would have a greater impact on the prediction. 

\begin{figure*}
    \centering
    \includegraphics[width=0.7\linewidth]{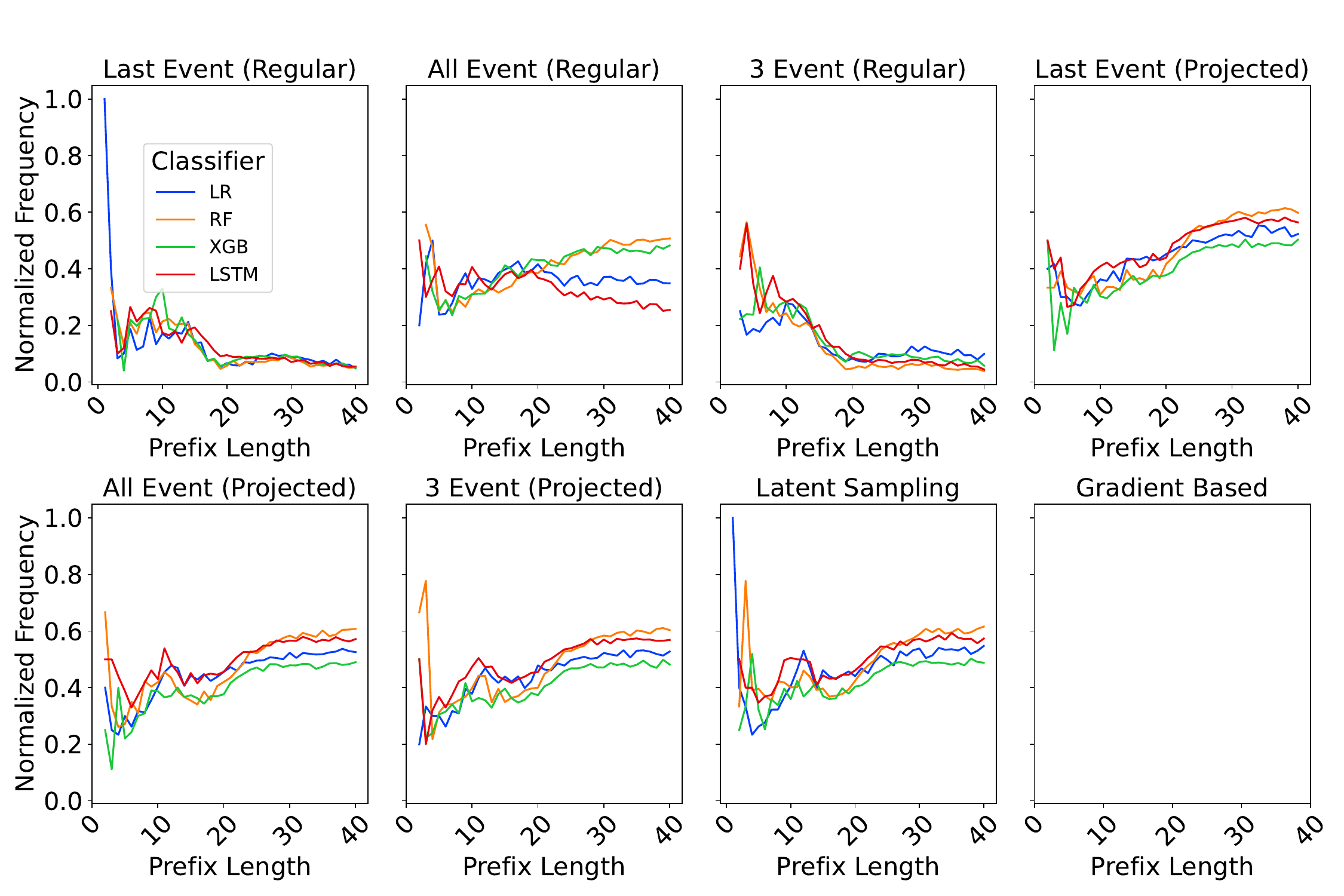}
    \caption{The normalized frequency of the \textbf{successful} adversarial examples (y-axis) per prefix length (x-axis), averaged over the event logs.}
    \label{fig: successrateperlength}
\end{figure*}

In Figure \ref{fig: successrateperlength}, we present the normalized frequency of successful adversarial attacks across different trace prefix lengths, averaged over the event logs, for various classifiers and attack methods. We focus on prefix length as it serves as an important factor in business process analysis, influencing the amount of information available to a predictive model at each point in a process. Longer prefixes typically provide a more comprehensive context, which can make models less sensitive to small perturbations, impacting the success rate of adversarial attacks. The results for the \emph{Last Event (Regular)} attack method indicate a negative relationship between trace length and attack success rate, suggesting that classifiers become less vulnerable to adversarial attacks when the activity in the last event of longer traces is permuted, compared to shorter ones. In contrast, for the \emph{projected}, \emph{latent sampling}, and \emph{gradient steps} attack methods, there is a clear positive relationship between prefix length and success rate, as the normalized frequency generally increases with longer prefixes. A similar trend is observed for the RF and XGB models under the \textit{All Event (Regular)} attack method, but this pattern does not hold for the LR and LSTM models.

\begin{table*}[ht]
\centering
\caption{Overview of the results}
\label{tab: results}
\resizebox{0.7\linewidth}{!}{%
\begin{tabular}{|c|clccccccc|}
\hline
\rowcolor[HTML]{EFEFEF} 
{\color[HTML]{000000} Event Log}                                          & {\color[HTML]{000000} $|\sigma|$} & \multicolumn{1}{c}{\cellcolor[HTML]{EFEFEF}{\color[HTML]{000000} \begin{tabular}[c]{@{}c@{}}Attack Type \\ and Strategy\end{tabular}}} & {\color[HTML]{000000} \begin{tabular}[c]{@{}c@{}}Success\\ Rate (\%)\end{tabular}} & {\color[HTML]{000000} \begin{tabular}[c]{@{}c@{}}Eucl. Dist.\\ (lat. space)\end{tabular}} & {\color[HTML]{000000} $L_{1}$} & {\color[HTML]{000000} $L_{2}$} & {\color[HTML]{000000} EMD} & {\color[HTML]{000000} DL Edit} & {\color[HTML]{000000} LCP} \\ \hline
\rowcolor[HTML]{FFFFFF} 
\cellcolor[HTML]{EFEFEF}{\color[HTML]{000000} }                           & {\color[HTML]{000000} 31}         & {\color[HTML]{000000} Last Event (Regular)}                                                                                            & 14                                                                                & 0.0013                                                                                    & 2.00                           & 1.41                           & 6.10                       & 1.00                           & {\color[HTML]{000000} 30}  \\
\rowcolor[HTML]{FFFFFF} 
\cellcolor[HTML]{EFEFEF}{\color[HTML]{000000} }                           & {\color[HTML]{000000} 31}         & {\color[HTML]{000000} All Event (Regular)}                                                                                             & 41                                                                                & 0.0201                                                                                    & 59.77                          & 7.70                           & 4.20                       & 29.40                          & {\color[HTML]{000000} 0}   \\
\rowcolor[HTML]{FFFFFF} 
\cellcolor[HTML]{EFEFEF}{\color[HTML]{000000} }                           & {\color[HTML]{000000} 31}         & {\color[HTML]{000000} 3 Event (Regular)}                                                                                               & 12                                                                                & 0.0071                                                                                    & 6.00                           & 2.45                           & 6.18                       & 3.00                           & {\color[HTML]{000000} 8}   \\
\rowcolor[HTML]{FFFFFF} 
\cellcolor[HTML]{EFEFEF}{\color[HTML]{000000} }                           & {\color[HTML]{000000} 39}         & {\color[HTML]{000000} Last Event (Projected)}                                                                                          & 66                                                                                & 0.0214                                                                                    & 54.16                          & 7.33                           & 15.09                      & 31.13                          & {\color[HTML]{000000} 4}   \\
\rowcolor[HTML]{FFFFFF} 
\cellcolor[HTML]{EFEFEF}{\color[HTML]{000000} }                           & {\color[HTML]{000000} 39}         & {\color[HTML]{000000} All Event (Projected)}                                                                                           & 66                                                                                & 0.0185                                                                                    & 54.17                          & 7.33                           & 15.06                      & 31.12                          & {\color[HTML]{000000} 4}   \\
\rowcolor[HTML]{FFFFFF} 
\cellcolor[HTML]{EFEFEF}{\color[HTML]{000000} }                           & {\color[HTML]{000000} 39}         & {\color[HTML]{000000} 3 Event (Projected)}                                                                                             & 66                                                                                & 0.0186                                                                                    & 54.15                          & 7.33                           & 15.07                      & 31.11                          & {\color[HTML]{000000} 4}   \\
\rowcolor[HTML]{FFFFFF} 
\cellcolor[HTML]{EFEFEF}{\color[HTML]{000000} }                           & {\color[HTML]{000000} 39}         & {\color[HTML]{000000} Latent Sampling}                                                                                                 & 66                                                                                & 0.0185                                                                                    & 54.18                          & 7.34                           & 15.11                      & 31.11                          & {\color[HTML]{000000} 4}   \\
\rowcolor[HTML]{FFFFFF} 
\multirow{-8}{*}{\cellcolor[HTML]{EFEFEF}{\color[HTML]{000000} BPIC2012}} & {\color[HTML]{000000} 35}         & {\color[HTML]{000000} Gradient Based}                                                                                                  & 53                                                                                & 0.0066                                                                                    & 26.66                          & 4.43                           & 11.12                      & 15.99                          & {\color[HTML]{000000} 6}   \\ \hline
\rowcolor[HTML]{FFFFFF} 
\cellcolor[HTML]{EFEFEF}{\color[HTML]{000000} }                           & {\color[HTML]{000000} 36}         & {\color[HTML]{000000} \textit{(averaged)}}                                                                                             & 48                                                                                & 0.0140                                                                                    & 38.89                          & 5.67                           & 10.99                      & 21.73                          & {\color[HTML]{000000} 8}   \\ \hline
\rowcolor[HTML]{FFFFFF} 
\cellcolor[HTML]{EFEFEF}{\color[HTML]{000000} }                           & {\color[HTML]{000000} 26}         & \cellcolor[HTML]{FFFFFF}{\color[HTML]{000000} Last Event (Regular)}                                                                    & 3                                                                                 & 0.0006                                                                                    & 2.00                           & 1.41                           & 32.26                      & 1.00                           & {\color[HTML]{000000} 25}  \\
\rowcolor[HTML]{FFFFFF} 
\cellcolor[HTML]{EFEFEF}{\color[HTML]{000000} }                           & {\color[HTML]{000000} 26}         & \cellcolor[HTML]{FFFFFF}{\color[HTML]{000000} All Event (Regular)}                                                                     & 38                                                                                & 0.0131                                                                                    & 52.07                          & 7.10                           & 74.11                      & 26.03                          & {\color[HTML]{000000} 0}   \\
\rowcolor[HTML]{FFFFFF} 
\cellcolor[HTML]{EFEFEF}{\color[HTML]{000000} }                           & {\color[HTML]{000000} 26}         & \cellcolor[HTML]{FFFFFF}{\color[HTML]{000000} 3 Event (Regular)}                                                                       & 6                                                                                 & 0.0087                                                                                    & 6.00                           & 2.45                           & 31.16                      & 3.00                           & {\color[HTML]{000000} 5}   \\
\rowcolor[HTML]{FFFFFF} 
\cellcolor[HTML]{EFEFEF}{\color[HTML]{000000} }                           & {\color[HTML]{000000} 12}         & \cellcolor[HTML]{FFFFFF}{\color[HTML]{000000} Last Event (Projected)}                                                                  & 38                                                                                & 0.0208                                                                                    & 33.16                          & 5.66                           & 13.75                      & 24.13                          & {\color[HTML]{000000} 2}   \\
\rowcolor[HTML]{FFFFFF} 
\cellcolor[HTML]{EFEFEF}{\color[HTML]{000000} }                           & {\color[HTML]{000000} 12}         & \cellcolor[HTML]{FFFFFF}{\color[HTML]{000000} All Event (Projected)}                                                                   & 37                                                                                & 0.0225                                                                                    & 32.50                          & 5.54                           & 13.96                      & 23.60                          & {\color[HTML]{000000} 2}   \\
\rowcolor[HTML]{FFFFFF} 
\cellcolor[HTML]{EFEFEF}{\color[HTML]{000000} }                           & {\color[HTML]{000000} 12}         & \cellcolor[HTML]{FFFFFF}{\color[HTML]{000000} 3 Event (Projected)}                                                                     & 37                                                                                & 0.0224                                                                                    & 32.49                          & 5.54                           & 13.96                      & 23.61                          & {\color[HTML]{000000} 2}   \\
\rowcolor[HTML]{FFFFFF} 
\cellcolor[HTML]{EFEFEF}{\color[HTML]{000000} }                           & {\color[HTML]{000000} 12}         & \cellcolor[HTML]{FFFFFF}{\color[HTML]{000000} Latent Sampling}                                                                         & 38                                                                                & 0.0222                                                                                    & 33.35                          & 5.69                           & 13.70                      & 24.27                          & {\color[HTML]{000000} 2}   \\
\rowcolor[HTML]{FFFFFF} 
\multirow{-8}{*}{\cellcolor[HTML]{EFEFEF}{\color[HTML]{000000} BPIC2015}} & {\color[HTML]{000000} 23}         & \cellcolor[HTML]{FFFFFF}{\color[HTML]{000000} Gradient Based}                                                                          & 4                                                                                 & 0.0060                                                                                    & 9.85                           & 1.90                           & 26.90                      & 6.54                           & {\color[HTML]{000000} 3}   \\ \hline
\rowcolor[HTML]{FFFFFF} 
\cellcolor[HTML]{EFEFEF}                                                  & 19                                & \textit{(averaged)}                                                                                                                    & 25                                                                                & 0.0145                                                                                    & 25.18                          & 4.41                           & 27.48                      & 16.52                          & 5                          \\ \hline
\rowcolor[HTML]{FFFFFF} 
\cellcolor[HTML]{EFEFEF}{\color[HTML]{000000} }                           & {\color[HTML]{000000} 13}         & \cellcolor[HTML]{FFFFFF}{\color[HTML]{000000} Last Event (Regular)}                                                                    & 24                                                                                & 0.0027                                                                                    & 2.00                           & 1.41                           & 1.11                       & 1.00                           & {\color[HTML]{000000} 12}  \\
\rowcolor[HTML]{FFFFFF} 
\cellcolor[HTML]{EFEFEF}{\color[HTML]{000000} }                           & {\color[HTML]{000000} 13}         & \cellcolor[HTML]{FFFFFF}{\color[HTML]{000000} All Event (Regular)}                                                                     & 35                                                                                & 0.0138                                                                                    & 23.68                          & 4.78                           & 2.15                       & 11.45                          & {\color[HTML]{000000} 0}   \\
\rowcolor[HTML]{FFFFFF} 
\cellcolor[HTML]{EFEFEF}{\color[HTML]{000000} }                           & {\color[HTML]{000000} 13}         & \cellcolor[HTML]{FFFFFF}{\color[HTML]{000000} 3 Event (Regular)}                                                                       & 28                                                                                & 0.0067                                                                                    & 6.00                           & 2.45                           & 1.12                       & 3.00                           & {\color[HTML]{000000} 3}   \\
\rowcolor[HTML]{FFFFFF} 
\cellcolor[HTML]{EFEFEF}{\color[HTML]{000000} }                           & {\color[HTML]{000000} 8}          & \cellcolor[HTML]{FFFFFF}{\color[HTML]{000000} Last Event (Projected)}                                                                  & 33                                                                                & 0.0121                                                                                    & 10.67                          & 2.72                           & 1.19                       & 7.39                           & {\color[HTML]{000000} 2}   \\
\rowcolor[HTML]{FFFFFF} 
\cellcolor[HTML]{EFEFEF}{\color[HTML]{000000} }                           & {\color[HTML]{000000} 7}          & \cellcolor[HTML]{FFFFFF}{\color[HTML]{000000} All Event (Projected)}                                                                   & 46                                                                                & 0.0092                                                                                    & 13.84                          & 3.59                           & 1.42                       & 9.60                           & {\color[HTML]{000000} 2}   \\
\rowcolor[HTML]{FFFFFF} 
\cellcolor[HTML]{EFEFEF}{\color[HTML]{000000} }                           & {\color[HTML]{000000} 7}          & \cellcolor[HTML]{FFFFFF}{\color[HTML]{000000} 3 Event (Projected)}                                                                     & 45                                                                                & 0.0093                                                                                    & 13.78                          & 3.57                           & 1.41                       & 9.56                           & {\color[HTML]{000000} 2}   \\
\rowcolor[HTML]{FFFFFF} 
\cellcolor[HTML]{EFEFEF}{\color[HTML]{000000} }                           & {\color[HTML]{000000} 7}          & \cellcolor[HTML]{FFFFFF}{\color[HTML]{000000} Latent Sampling}                                                                         & 48                                                                                & 0.0098                                                                                    & 14.32                          & 3.72                           & 1.44                       & 9.91                           & {\color[HTML]{000000} 2}   \\
\rowcolor[HTML]{FFFFFF} 
\multirow{-8}{*}{\cellcolor[HTML]{EFEFEF}{\color[HTML]{000000} SEPSIS}}   & {\color[HTML]{000000} 12}         & \cellcolor[HTML]{FFFFFF}{\color[HTML]{000000} Gradient Based}                                                                          & 45                                                                                & 0.0048                                                                                    & 7.42                           & 2.42                           & 1.14                       & 5.30                           & {\color[HTML]{000000} 6}   \\ \hline
\rowcolor[HTML]{FFFFFF} 
\cellcolor[HTML]{EFEFEF}{\color[HTML]{000000} }                           & {\color[HTML]{000000} 10}         & {\color[HTML]{000000} \textit{(averaged)}}                                                                                             & 38                                                                                & 0.0086                                                                                    & 11.46                          & 3.08                           & 1.37                       & 7.15                           & {\color[HTML]{000000} 4}   \\ \hline
\end{tabular}%
}
\end{table*}

\begin{figure*}
    \centering
\includegraphics[width=1\linewidth]{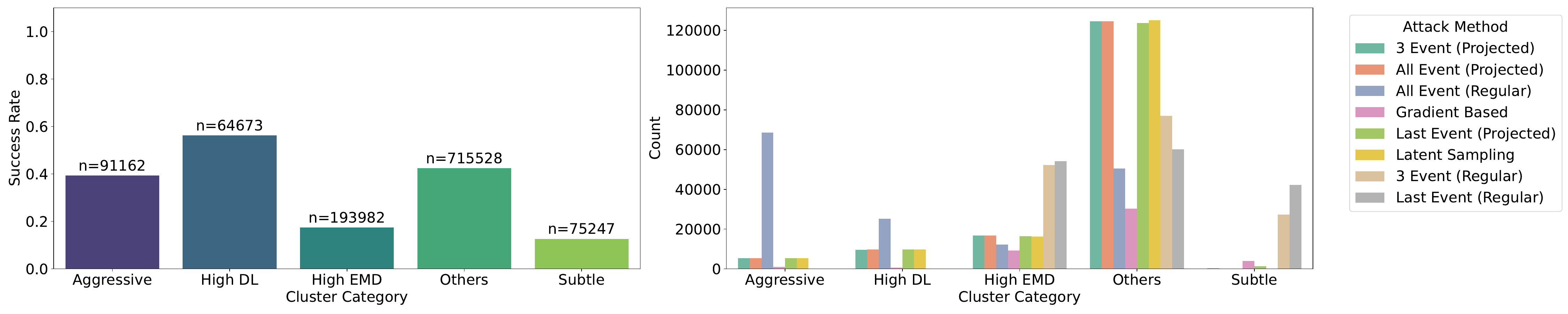}
    \caption{The success rate percentage and absolute number of traces per cluster category (left), and the number of traces per cluster category and attack method.}
    \label{fig: attacktypeanalysis}
\end{figure*}

\subsection{Results per Event Log Group}

In Table \ref{tab: results}, we present the results for the three groups of event logs, averaged per group. 

The Euclidean distance represents the distance in the latent space, where lower values indicate that adversarial examples are more perceptibly indistinguishable from the original instances. Interestingly, the Euclidean distance in the latent space for regular attack strategies (apart from the \emph{All Event (Regular)}) is lower than those for projected strategies, such as \emph{Last Event (Projected)}, \emph{All Event (Projected)}, and \emph{3 Event (Projected)}. This is counterintuitive, as one might expect that projection-based attacks, designed to be close in the latent manifold, would yield adversarial examples that are actually closer. This contradicts previous findings in \cite{stevens2023manifold}. 

The \textit{EMD} captures the distance between the original prefix and its adversarial example to provide a meaningful measure of similarity between probability distributions, which, in the context of adversarial examples, reflects how \textit{close} the generated adversarial example is to the original prefix trace in terms of its underlying data distribution. The \textit{EMD} serves as a measure of distance in the input space, with lower values indicating adversarial examples that are more perceptually indistinguishable from the original instances. In adversarial machine learning, this is crucial because we want adversarial examples to remain realistic and perceptually similar to the original input while still achieving the goal of deceiving the model. 

Another interesting remark is that, in general, the \textit{EMD} distances of the adversarial examples are lower for the \emph{projected} attacking strategy compared to the \emph{regular} attacking strategies. For the event log group \emph{BPIC2012}, we can see that the \emph{regular} attacking strategy methods have the lowest average distance in the latent space (Euclidean distance), but also the highest \textit{EMD} value of the adversarial examples after decoding. To understand this intuitively, we need to understand how the decoder functions. First, minimizing distances in the latent space encourages adversarial examples to stay close to the original representations at an encoded level. However, this does not guarantee that the decoded outputs will remain perceptually close to the original inputs. This is because the decoder is trained to reconstruct training instances, and the sensitivity of the decoder to changes in the latent space may vary based on the characteristics of the dataset it was trained on. As a result, even small shifts in latent space can lead to disproportionately large changes in the decoded output if the decoder is highly sensitive. 

\subsection{Cluster Profiles of Attacks}\label{subsec: clustering}
\begin{table}[ht]
\centering
\caption{Cluster profiles (overview)}
\label{tab: clusterprofiles}
\resizebox{\columnwidth}{!}{%
\begin{tabular}{|
>{\columncolor[HTML]{EFEFEF}}c |
>{\columncolor[HTML]{FFFFFF}}c 
>{\columncolor[HTML]{FFFFFF}}c 
>{\columncolor[HTML]{FFFFFF}}c 
>{\columncolor[HTML]{FFFFFF}}c 
>{\columncolor[HTML]{FFFFFF}}c |}
\hline
{\color[HTML]{000000} \begin{tabular}[c]{@{}c@{}}Cluster \\ profile\end{tabular}}      & \cellcolor[HTML]{EFEFEF}{\color[HTML]{000000} DL Edit}                           & \cellcolor[HTML]{EFEFEF}{\color[HTML]{000000} EMD}                               & \cellcolor[HTML]{EFEFEF}{\color[HTML]{000000} Impact}                              & \cellcolor[HTML]{EFEFEF}{\color[HTML]{000000} Typical attack}                               & \cellcolor[HTML]{EFEFEF}{\color[HTML]{000000} Perceptible?} \\ \hline
{\color[HTML]{000000} Aggressive}                                                      & {\color[HTML]{000000} High}                                                      & \cellcolor[HTML]{FFFFFF}{\color[HTML]{000000} High}                              & {\color[HTML]{000000} Successful}                                                  & \cellcolor[HTML]{FFFFFF}{\color[HTML]{000000} All event}                                    & \cellcolor[HTML]{FFFFFF}{\color[HTML]{000000} $\checkmark$} \\ \hline
{\color[HTML]{000000} \begin{tabular}[c]{@{}c@{}}Sequence\\ perturbation\end{tabular}} & {\color[HTML]{000000} High}                                                      & {\color[HTML]{000000} \begin{tabular}[c]{@{}c@{}}Low to\\ moderate\end{tabular}} & {\color[HTML]{000000} \begin{tabular}[c]{@{}c@{}}Highly\\ successful\end{tabular}} & {\color[HTML]{000000} All event}                                                            & {\color[HTML]{000000} $\checkmark$}                         \\ \hline
{\color[HTML]{000000} \begin{tabular}[c]{@{}c@{}}Distribution \\ shift\end{tabular}}   & {\color[HTML]{000000} \begin{tabular}[c]{@{}c@{}}Low to\\ moderate\end{tabular}} & {\color[HTML]{000000} High}                                                      & {\color[HTML]{000000} \begin{tabular}[c]{@{}c@{}}Not\\ successful\end{tabular}}    & {\color[HTML]{000000} \begin{tabular}[c]{@{}c@{}}Last event, \\ 3 event\end{tabular}}       & {\color[HTML]{000000} $\checkmark$}                         \\ \hline
{\color[HTML]{000000} Others}                                                          & {\color[HTML]{000000} \begin{tabular}[c]{@{}c@{}}Low to\\ moderate\end{tabular}} & {\color[HTML]{000000} \begin{tabular}[c]{@{}c@{}}Low to\\ moderate\end{tabular}} & {\color[HTML]{000000} Successful}                                                  & {\color[HTML]{000000} \begin{tabular}[c]{@{}c@{}}Projected,\\ latent sampling\end{tabular}} & {\color[HTML]{000000} Sometimes}                            \\ \hline
{\color[HTML]{000000} Subtle}                                                          & {\color[HTML]{000000} Low}                                                       & {\color[HTML]{000000} Low}                                                       & {\color[HTML]{000000} \begin{tabular}[c]{@{}c@{}}Not\\ successful\end{tabular}}    & {\color[HTML]{000000} \begin{tabular}[c]{@{}c@{}}Last event, \\ 3 event\end{tabular}}       & {\color[HTML]{000000} -}                                    \\ \hline
\end{tabular}%
}
\end{table}
Instead of relying solely on statistical analysis, we define these clusters as profiles that capture common patterns in the nature of adversarial changes. To group attacks by their unique characteristics, we apply thresholds derived from quartile values, calculated after normalizing both the \textit{EMD} and \textit{DL Edit} scores according to the prefix length. This normalization ensures comparability across traces of varying lengths. The quartile-based thresholds provide a straightforward yet effective method for distinguishing between attack cluster profiles.

The resulting profiles categorize attacks into four primary groups: \textbf{Subtle}, \textbf{Aggressive}, \textbf{Sequence Perturbations}, \textbf{Distribution Shift}. Attacks that do not exhibit sufficiently \textit{distinctive} characteristics are classified as \textbf{Others}. Below, we detail the criteria used to define and classify each profile.\\

\noindent \textbf{Aggressive}
Aggressive attacks are characterized by significant changes to the trace. Specifically, an attack is classified as aggressive if the \textit{DL Edit} distance and \textit{EMD distance} fall within the third quartile, indicating extensive changes to the trace and its overall distribution. While these attacks achieve moderate success rates (around 40\%), their broad and noticeable changes to the trace make them relatively easier to detect. It is clear to see that the \textit{All Event (Regular)} category is predominantly represented within this cluster, as these attacks indeed target all the events in a trace.\\

\noindent \textbf{Subtle Attacks}
Subtle attacks are those that make only minimal changes to the original trace. An attack is considered subtle if both the \textit{DL Edit} distance and the \textit{EMD} value fall within the first quartile. These minimal changes reflect the conservative nature of such strategies and their focus on remaining undetected by only performing small perturbations that are less likely to be detected in the event log. As expected, the \textit{Last Event (Regular)} and \textit{3 Event (Regular)} profiles are best represented by this classification. These attacks are the least successful attacks, with a success rate of below 20 \%.\\

\noindent \textbf{Sequence Perturbations.}
Sequence Perturbations are changes that specifically target the order, dependencies, and relationships between events in the trace. Attacks with a high \textit{DL Edit} are adversarial examples with a \textit{DL Edit} distance in the third quartile, but also with an \textit{EMD value} that is lower than the median value. This kind of attack primarily targets the sequence order of events without altering the underlying distribution of event types, and they are highly successful, with a success rate of up to 60 percent. This group is mostly represented by the \textit{All Event (Regular)} attack.\\

\noindent \textbf{Distribution Shift.} Distribution shift attacks are attacks that made changes to the trace that, while not necessarily altering individual events in an obvious or direct way, lead to a significant shift in the overall structure or distribution of events.
Attacks with a high \textit{EMD} value are adversarial examples with an \textit{EMD} distance in the third quartile. These attacks are not that successful, with a success rate close to 20 percent. This group is mostly represented by the \textit{3 Event (Regular)} attack and the \textit{Last Event (Regular)} attack.\\

\noindent \textbf{Others.} These attacks can be characterized by a combination of small changes in sequence structure (as indicated by a low to moderate DL) and slight shifts in event distribution (reflected by a low to moderate EMD), yet they are still successful. Although these attacks modify the sequence structure and event distribution of the trace, they are subtle enough to the extent that they would not be easily detectable. Despite these modest changes, the attacks remain successful, with a success rate of up to 50\%. \\

In conclusion, the \emph{Others} cluster represents the most desirable attack profile, as these attacks generated adversarial examples with a low to moderate \textit{EMD} distance and low to moderate \textit{DL Edit} value, while maintaining high success rates. The fact that these attacks introduce more or less \textit{small} changes in both sequence structure and event distribution, yet remain effective, suggests they exploit vulnerabilities that are not easily detected. The attack methods most representative of this category include the \textit{Last Event (Projected)}, the \textit{All Event (Projected)}, the \textit{3 Event (Projected)} and the \textit{Latent Sampling} attacks. These methods showcase how moderate perturbations to the trace can yield successful results without major alterations to the \textit{original process flow}.
This showcases the need for projected and latent sampling-based approaches, as they can be effective without needing any particular engineering which resorts to strong insights into the sequential information of the processes.

\section{Related Work}\label{sec: relatedwork}

\subsection{Predictive Process Monitoring}
Predictive process monitoring is an important field that involves monitoring and analyzing process data extracted from business information
systems. In recent years, researchers have proposed various solutions to tackle different prediction tasks, such as predicting
the most probable next event or suffix of a case \cite{tax2017predictive}, estimating
the remaining time \cite{van2008cycle}, or predicting the outcome \cite{DBLP:journals/tkdd/TeinemaaDRM19}. For
a more comprehensive overview of the field, we refer to the systematic literature review of Neu et al. \cite{neu2022systematic}.

Recent works have already issued the lack of reliability of deep learning models in the context of predictive process
monitoring \cite{stevens2024explainability, DBLP:rizzi} with issues such as the compromised faithfulness of post-hoc explanations \cite{stevens2024explainability}. Incremental adaptations to predictive models have also been suggested to address prediction stability over time \cite{DBLP:rizzi}.

\subsection{Adversarial Machine Learning}
    
Deep neural networks are powerful tools for learning complex tasks, but their adoption in high-stake decision-making is often limited due to their lack of robustness against adversarial examples \cite{DBLP:journals/corr/SzegedyZSBEGF13, DBLP:journals/corr/GoodfellowSS14}. Adversarial Machine Learning (AML) aims to test and enhance this robustness by generating adversarial examples designed to deceive the algorithms. Adversarial attacks on time series prediction models have already been investigated \cite{wu2022small}.

In our previous work, we examined the vulnerability of OOPPM models and their explainability methods to adversarial attacks, highlighting risks to decision-making accuracy \cite{stevens2022assessing}. These attacks are (naively) engineered to fool models into producing incorrect predictions and pose a risk to the accuracy of high-stakes decision-making processes. In more of our recent work \cite{stevens2023manifold}, the idea was to explore these adversarial attacks as a means to enhance robustness against such threats. 

The field of AML is also focused on improving robustness model and generalization \cite{DBLP:conf/cvpr/Stutz0S19, DBLP:taymouripredictive, DBLP:taymourideep, DBLP:conf/bpm/VenkateswaranMI21, DBLP:conf/caise/KappelJ23}. Defense mechanisms against adversarial examples and taxonomies of such examples have been proposed \cite{DBLP:journals/tnn/YuanHZL19}. Generative Adversarial Networks (GANs) have also been adapted for robust predictive modeling, with innovations like the closed-loop adversarial training using Encoder-Decoder (ED) architectures \cite{DBLP:taymourideep, DBLP:taymouripredictive} to improve the suffix and remaining time prediction of event sequences. Next, robust predictive models that adapt to spurious correlations and varying data distributions have demonstrated improved performance in real-world event log evaluations \cite{DBLP:conf/bpm/VenkateswaranMI21}. Simple noise-based transformations have been proposed to enhance event log data for next activity prediction \cite{DBLP:conf/caise/KappelJ23}.

\subsection{Manifold Learning}
Manifold learning aims to uncover the underlying structure of high-dimensional data, particularly useful in multivariate variable-length time series \cite{ho2015manifold}. Although early theories attributed model vulnerability to rare cases or the linear nature of DNNs \cite{DBLP:journals/corr/SzegedyZSBEGF13, DBLP:journals/corr/GoodfellowSS14}, recent works emphasize the importance of generating realistic, on-manifold adversarial examples \cite{zhao2018generating, schott2019towards}. In predictive process monitoring, VAEs have been used to learn latent space data representations of process data \cite{stevens2023manifold, stevens2024generating}. 
\section{Conclusion}\label{sec: Robustconclusion}

In this paper, we compared and evaluated four distinct adversarial attack strategies, resulting in eight different attacking methods, all specifically tailored to the process-based analysis domain. Our work addresses critical challenges in OOPPM by thoroughly evaluating model vulnerabilities to adversarial attacks and generating realistic adversarial examples. We benchmarked a variety of black-box and white-box attacks, highlighting that even small perturbations, whether in the original data space or the latent space, can significantly impact the performance of predictive models. This underscores the susceptibility of predictive models in PPM to adversarial examples, which can undermine their reliability in real-world applications.

One of the key insights from our analysis is that adversarial robustness in PPM cannot be approached in the same way as in traditional domains. Business processes have unique characteristics—such as the sequential nature of events, their dependencies, and the historical context of activities—that require specialized adversarial attack methods and defense mechanisms. Our introduction of three novel attacks, including a data-aware approach that respects the historical positioning of activities and two latent-space-based attacks, illustrates the importance of designing attacks that account for these intricacies. By doing so, we provide a more nuanced understanding of how adversarial examples manifest in business processes and how predictive models respond to them.

Excluding the fact that adversarial examples were generated only from instances that the original model initially predicted correctly, we can conclude that all the attack generation methods, except for the gradient-based approach, are model-agnostic methods. This means the attacks generated from these can be stored in a new test set useful for evaluating and comparing robustness across various models. The gradient-based attacks, which are generated with the help of using a trained OOPPM (LSTM) model, can also be collected into a robustness test set. However, this should not be used to compare the robustness of that particular model to other models, since the attacks are explicitly made based on its gradients.

Moreover, our study emphasizes the broader challenge of ensuring trust in PPM models. This trust can only be established if the models are robust against adversarial examples, which could otherwise lead to incorrect predictions and costly decision-making errors. To address this, we introduced a comprehensive robustness evaluation framework. This framework facilitates the generation of diverse adversarial examples, allowing models to be tested and trained under adversarial conditions. 

Future work could be to identify which types of traces (e.g., common patterns, rare outliers, or specific structural characteristics) are most prone to adversarial perturbation. Additionally, examining how process models change before and after attacks can provide valuable insights into the nature of adversarial examples and inform strategies for mitigating their impact. Another important direction is investigating whether models should be trained separately on each attack type or on a mixture of multiple attacks to enhance robustness. Visualizing the structure of the latent manifold, particularly through techniques such as convex hull visualization, could further deepen our understanding of how adversarial examples navigate the latent space and interact with the underlying data.

\bibliographystyle{splncs04}
\bibliography{references.bib}

\end{document}